\documentclass[runningheads]{llncs}
\usepackage{graphicx}
\usepackage{amsmath,amssymb} 
\usepackage{color}

\usepackage{multirow,array}
\usepackage[table]{xcolor}
\usepackage{subfigure}
\usepackage[noend]{algorithm2e} 
\usepackage{dsfont}
\usepackage{balance} 
\usepackage{enumitem}
\usepackage{cite}
\usepackage{mathrsfs}
\graphicspath{ {./images/} }
\usepackage[pagebackref=true,colorlinks]{hyperref}

\DeclareMathOperator{\Lset}{\mathcal{L}}
\begin{document}
\pagestyle{headings}
\mainmatter

\def\ACCV20SubNumber{373}  

\title{Leveraging Tacit Information Embedded in CNN Layers for Visual Tracking} 
\titlerunning{Tacit Information in CNN Layers for Tracking}
%
\author{Kourosh Meshgi\inst{1} \and
Maryam Sadat Mirzaei\inst{1} \and
Shigeyuki Oba\inst{2}} 
\authorrunning{K. Meshgi et al.}
%
\institute{RIKEN Center for Advanced Intelligence Project (AIP), Tokyo, Japan \and
Graduate School of Informatics, Kyoto University, Kyoto, Japan
}

\maketitle

\begin{abstract}
Different layers in CNNs provide not only different levels of abstraction for describing the objects in the input but also encode various implicit information about them. The activation patterns of different features contain valuable information about the stream of incoming images: spatial relations, temporal patterns, and co-occurrence of spatial and spatiotemporal (ST) features. The studies in visual tracking literature, so far, utilized only one of the CNN layers, a pre-fixed combination of them, or an ensemble of trackers built upon individual layers. 
In this study, we employ an adaptive combination of several CNN layers in a single DCF tracker to address variations of the target appearances and propose the use of style statistics on both spatial and temporal properties of the target, directly extracted from CNN layers for visual tracking.
Experiments demonstrate that using the additional implicit data of CNNs significantly improves the performance of the tracker. Results demonstrate the effectiveness of using style similarity and activation consistency regularization in improving its localization and scale accuracy.
\end{abstract}

\section{Introduction}
\begin{figure}[!t]
\centering
\includegraphics[width=0.6\linewidth]{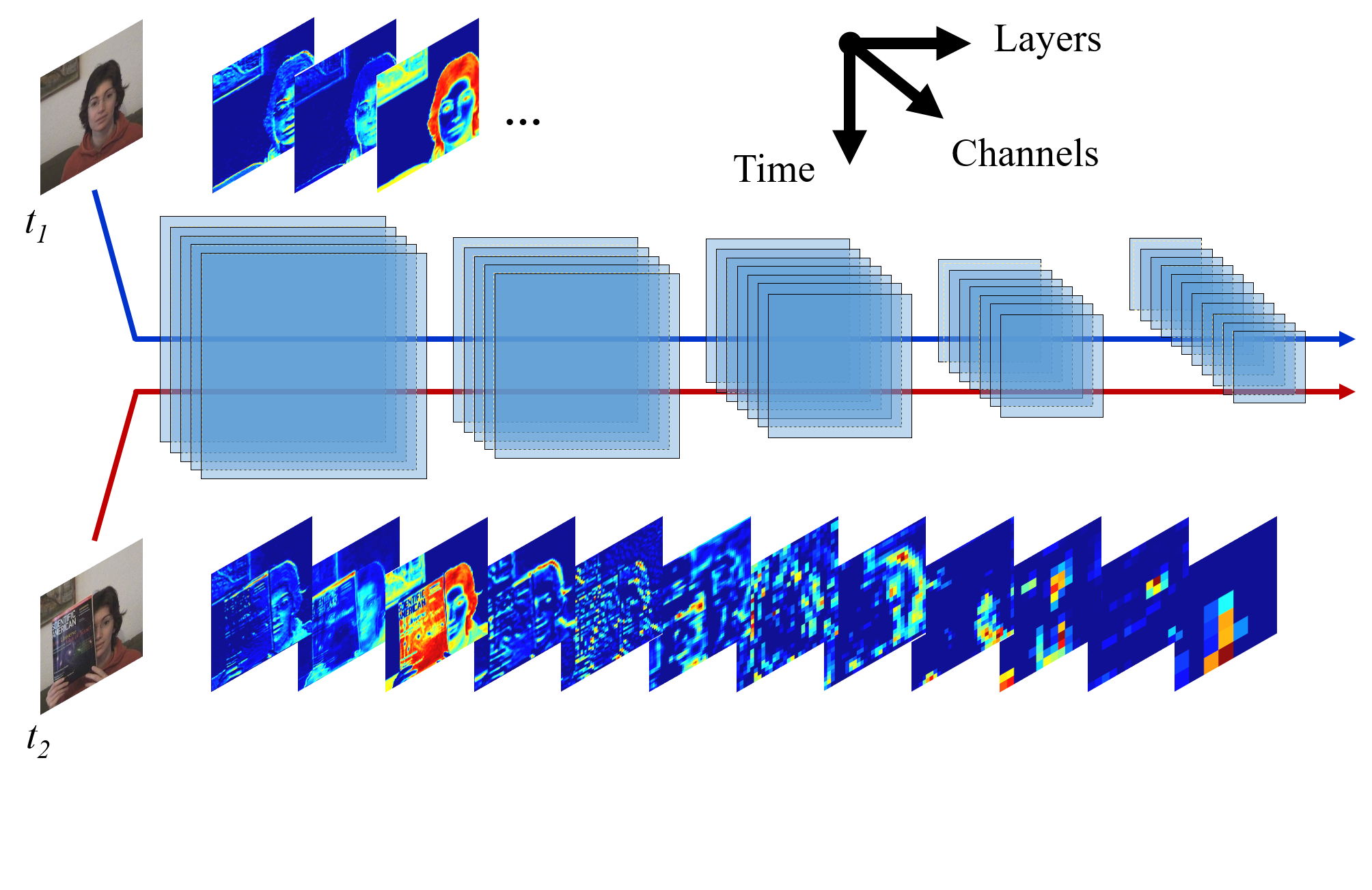} 
\begin{minipage}[b]{0.38\linewidth}
\centering
\subfigure[mixture of layers]{\includegraphics[width=0.7\linewidth]{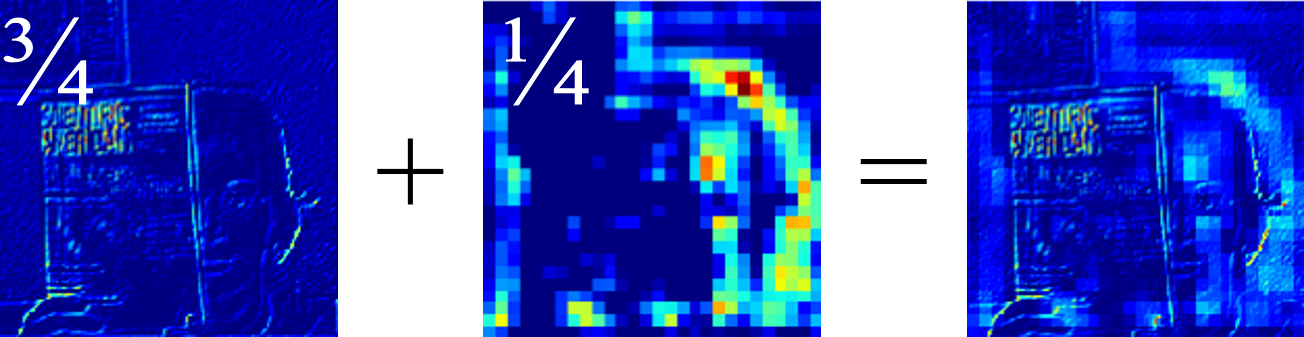}} \vspace{-0.3 cm}\\
\subfigure[spatial reg.]{\includegraphics[width=0.7\linewidth]{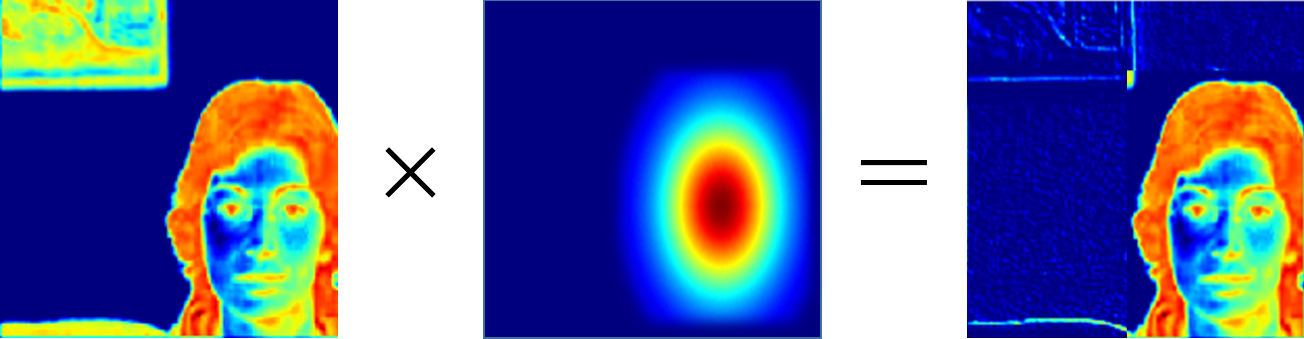}}\vspace{-0.3 cm}\\
\subfigure[comparing styles]{\includegraphics[width=0.7\linewidth]{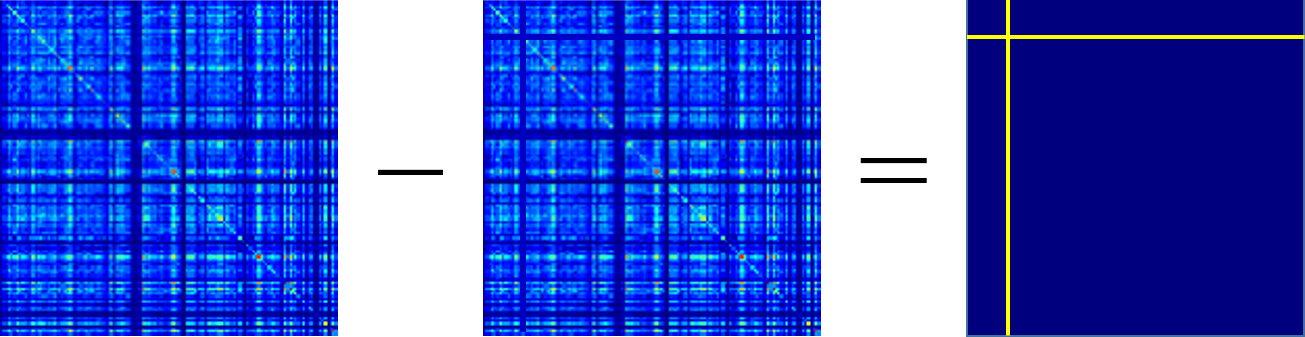}}\vspace{-0.3 cm}\\
\subfigure[temporal reg.]{\includegraphics[width=0.7\linewidth]{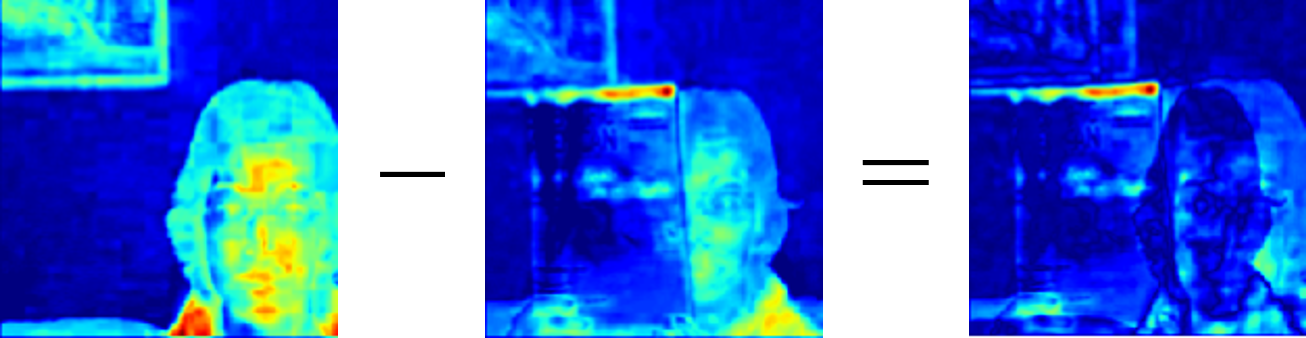}}\vspace{-0.3 cm}
\end{minipage}
\caption{When presented with a sequence of images, CNN neurons are activated in a particular way, involving information about the spatial, semantic, style and transformations of the target. \textbf{(a)} Combining information from different layers balances the amount of spatial and semantic information, \textbf{(b)} spatial weighting of the response would discard most of the background discratction, \textbf{(c)} changes of the co-activations of neurons (measured by Gram matrices) for different subsequent images indicate style changes of the target (the plot is exaggeratedly enhanced for visibility), and \textbf{(d)} changes in the activations of the neurons themselves signals the appearance transformations and pose changes (in shallower layers) or alteration of semantic contents. }
\label{fig:cnn_layers}
\end{figure}

Discovering new architectures for deep learning and analyzing their properties, have resulted in a rapid expansion in computer vision, along with other domains. Among these architectures, convolutional neural networks (CNNs) have played a critical role to capture the statistics and semantics of natural images. CNNs are widely used in different computer vision applications since they are able to effectively learn complicated mappings while using minimal domain knowledge \cite{DEEPTRACK}.
Deep learning has been introduced to visual tracking in different forms, mostly to provide features for established trackers based on correlation filters \cite{DeepSRDCF}, particle filters \cite{wang2015video,zhang2017multi} and detector-based trackers \cite{CNN-SVM}. Although deep features extracted by fully-connected layers of CNN are shown to be adequately generic to be used for a variety of computer vision tasks, including tracking \cite{sharif2014cnn}. Further studies revealed that convolutional layers are even more discriminative, semantically meaningful and capable of learning structural information \cite{cimpoi2014deep}. However, the direct use of CNNs to perform tracking is complicated because of the need to re-train the classifier with the stream of target appearances during tracking, the diffusion of background information into template \cite{danelljan2015learning}, and ``catastrophic forgetting'' of previous appearance in the face of extreme deformations and occlusions \cite{TCNN}.

Early studies in the use of deep learning in tracking utilized features from autoencoders \cite{DLT,AE-ENS} and fully-connected layers of pre-trained (CNN-based) object detector \cite{fan2010human}, but later the layers were used to serve as features balancing the abstraction level needed to localize the target \cite{CF2}, provide ST relationship between the target and its background \cite{CNT}, combine spatial and temporal features \cite{UCT,YCNN}, and generate probability maps for the tracker \cite{SO-DLT}.
%
Recently trackers employ other deep learning approaches such as R-CNN for tracking-by-segmentation \cite{drayer2016object}, Siamese Networks for template similarity measuring \cite{SINT,SIAMESEFC,li2018high,li2019siamrpn++}, GANs to augment positive samples \cite{SINTpp,VITAL}, and CNN-embeddings for self-supervised image coloring and tracking \cite{Coloring}. However, the tacit information in pre-trained CNNs including information between layers, within layers, and activation patterns across the time axis are underutilized (Fig. \ref{fig:cnn_layers}). 

Different layers of CNNs provide different levels of abstraction \cite{zeiler2014visualizing}, and it is known that using multiple layers of CNNs can benefit other tasks such as image classification \cite{liu2015treasure}. Such information was used as a coarse-to-fine sifting mechanism in \cite{CF2}, as a tree-like pre-defined combination of layers with fixed weights and selective updating \cite{TCNN}, as the features for different trackers in an ensemble tracker \cite{HDTstar}, or in a summation over all layers as uniform features \cite{CCOT}. However, direct adaptive fusion of these features in a single tracker that can address different object appearances is still missing in the literature. 

CNN stores information not only between layers but within layers in different channels, each representing a different filter. These filters may respond to different visual stimuli. The shallower layers have a Gabor-like filter response \cite{bovik1990multichannel} whereas in the deeper layers, they respond to angles, color patterns, simple shapes, and gradually highly complex stimuli like faces \cite{zeiler2014visualizing}. The co-occurring patterns within a layer, activate two or more different channels of the layer. Such co-incidental activations are often called style in the context of neural style transfer (NST), and different approaches are proposed to measure the similarity between two styles \cite{gatys2016image,johnson2016perceptual}. The loss functions for NST problem can serve as the similarity index for two objects (e.g., in the context of image retrieval \cite{matsuo2016cnn}). 

Most of the current CNN-based techniques use architectures with 2D convolutions to achieve different invariances to the variations of the images. Meanwhile, the invariance to transformations in time is of paramount importance for video analysis \cite{varol2018long}. Modeling temporal information in CNNs has been tackled by applying CNNs on optical flow images \cite{gkioxari2015finding}, reformulating R-CNNs to exploit the temporal context of the images \cite{chao2018rethinking} or by the use of separate information pathways for spatial and temporal pathways \cite{simonyan2014two,zhu2017end}. Motion-based CNNs typically outperform CNN representations learned from images for tasks dealing with dynamic target, e.g. action recognition \cite{varol2018long}. In these approaches, a CNN is applied on 3-channel optical flow image \cite{dosovitskiy2015flownet}, and different layers of such network provide different variances toward speed and the direction of the target's motion \cite{feichtenhofer2018have}. In visual tracking, deep motion features provide promising results \cite{gladh2016deep}. However, this requires the tracker to fuse the information from two different CNN networks (temporal+spatial) \cite{zhu2017end}, and their inconsistency hinders a meaningful layer-wise fusion and only the last layers of temporal CNN are used for tracking \cite{gladh2016deep}. 

\begin{figure*}[!t]
\includegraphics[width=1\linewidth]{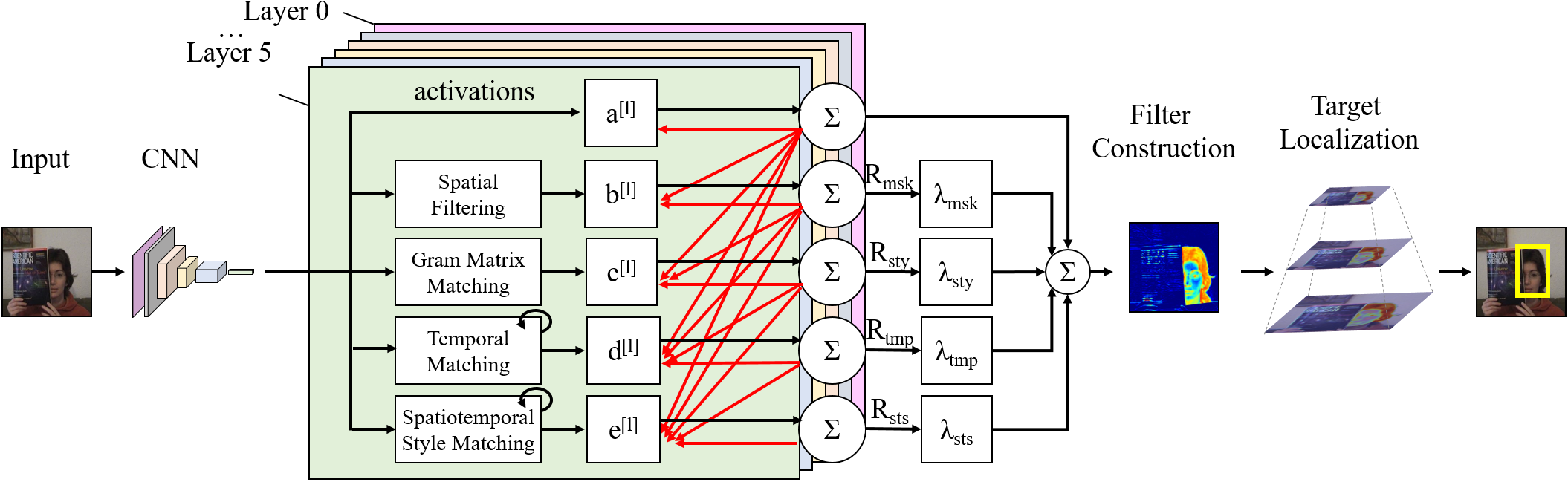}
\caption{Schematic of the proposed tracker. Given the input image, the activations of different layers of CNN are processed independently and its spatial irregularities, style mismatches, temporal inconsistencies and ST pattern changes compared to the template are calculated. These errors are then adaptively fused with that of other layers to form the regularization terms $R_x$. The final filter is then constructed and used in the baseline DCF tracker that uses multi-scale matching to find the position and scale of the target in the input image. The weights of different error terms are then updated in reverse proportion of their contribution in total error of each level.}
\label{fig:schematic}
\end{figure*}

\noindent\textbf{Contributions:}
We propose an adaptive fusion of different CNN layers in a tracker to combine high spatial resolution of earlier layers for precise localization, and semantic information of deeper layers to handle large appearance changes and alleviate model drift. We utilize the tacit information between each layer's channels at several timepoints, i.e., the style, to guide the tracker. To our best knowledge, this is the first use of within-layer information of CNNs especially spatial and ST co-activations for tracking. We also introduced temporal constraint helps to better preserve target's temporal consistency, suppress jitter, and promote scale adaptation.
\begin{itemize}[leftmargin=*,noitemsep]
\item[(i)] We propose an intuitive adaptive weight adjustment to tune the effect of components (spatial, background avoiding, co-incidental, temporal, and ST) both within and between CNN layers. Temporal and style regs are typically complementary: target changes are punished by a reduction in activation and style similarity, big changes are punished by spatial reg and style keep track of the target. Employing multiple layers not only gives different realization of details-semantics trade-off \cite{CF2}, but also provides richer statistics compared to one layer. We incorporate different regularization terms on CNN layers using only one feed-forward pass on an arbitrary pre-trained CNN, with no change to its architecture or CNN block design (e.g. ST block as in \cite{zhu2019stresnet_cf}) and no additional computation (e.g., compute optical flow).
\item[(ii)] We introduce a Gram-based style matching in our tracker to capture style alterations of the target. The style matching exploits the co-activation patterns of the layers and improves the localization accuracy of the target and provides complementary information to the baseline which relies on spatial matching.
\item[(iii)] We introduce the temporal coherence regularization to the tracker by monitoring activations of each layer through the course of tracking, to enhance tracker's movement speed and direction invariance, adaptation to different degrees of changes in the target appearance, stability, and scale tuning. 
\item[(iv)] Our system is tested on various public datasets (OTB50 \& 100, LaSOT, VOT2015 \& 2018, and UAV123), and considering the simplicity of our baseline (SRDCF \cite{danelljan2015learning}) we obtained results on par with many sophisticated trackers. The results shed light on the hidden information within CNN layers, that is the goal of this paper. The layer-fusion, style-matching, and temporal regularization component of the proposed tracker is further shown to advance the baseline significantly and outperformed state-of-the-art trackers.
\end{itemize}

It should be noted that our proposed method differs with HDT \cite{HDTstar}, CCOT \cite{danelljan2016beyond} and ECO \cite{danelljan2017eco} that also integrates the multi-layer deep features by considering them in the continuous domain. Here, we employed a pre-trained CNN for object detection task, simplified the need to deal with different dimensions of convolutional layers by maintaining the consistency between layers while isolating them to make an ensemble of features (rather than an ensemble of trackers in \cite{HDTstar}). Additionally, we proposed style and temporal regularization that can be incorporated into Conjugate Gradient optimization of C-COT \cite{CCOT}, ADMM optimization of BACF \cite{galoogahi2017learning} and Gauss-Newton optimization of ECO \cite{danelljan2017eco}.

\section{Method}
We propose a tracker that adaptively fuse different convolutional layers of a pre-trained CNN into a DCF tracker. The adaptation is inspired by weight tuning of the different tracker components as well as the scale-aware update of \cite{danelljan2014accurate}. As Figure \ref{fig:schematic} illustrates, we incorporated spatial regularization (a multi-layer extension of \cite{DeepSRDCF}), co-incidental regularization (which matches style between the candidate patch and historical templates), temporal regularization (that ensures a smooth alteration of the activations in normal condition), and ST regularization that captures the change patterns of spatial features.

\subsection{Discriminative Correlation Filter Tracker}
The DCF framework utilizes the properties of circular correlation to efficiently train and apply a classifier in a sliding window fashion \cite{henriques2012exploiting}. The resulting classifier is a correlation filter applied to the image/feature channels, similar to conv layers in a CNN. Different from CNNs, the DCF is trained by solving a linear least-squares problem using Fast Fourier Transform (FFT) effectively.

A set of example patches $x_\tau$ are sampled at each frame $\tau = 1,\ldots,t$ to train the discriminative correlation filer $f_t$, where $t$ denotes the current frame. The patches are all of the same size (conveniently, the input size of the CNN) centered at the estimated target location in each frame. We define feature $x^k_\tau$ as the output of channel $k$ at a convolutional layer in the CNN. With this notion, the tracking problem is reduced to learn a filter $f^k_t$ for each channel $k$, that minimizes the $L^2$-error between the responses $S_{f_t}$ on samples $x_\tau$ and the desired filter form $y_k$:
\begin{equation}
\epsilon = \sum_{\tau=1}^t \alpha_\tau ||S (f_t , x_\tau) - y_\tau||^2 + \lambda ||f_t||^2
\label{eq:dcf_loss}
\end{equation}
where $S(f_t,x_\tau) = f_t \star x_\tau$ in which the $\star$ denotes circular correlation generalized to multichannel signals by computing inner products. The desired correlation output $y_\tau$ is set to a Gaussian function with the peak placed at the target center location \cite{bolme2010visual}. A weight parameter $\lambda$ controls the impact of the filter size regularization term, while the weights $\alpha_\tau$ determine the impact of each sample.

To find an approximate solution of eq\eqref{eq:dcf_loss}, we use the online update rule of \cite{danelljan2014accurate}. At frame $t$, the numerator $g_t$ and denominator $\hat{h}_t$ of the discrete Fourier transformed (DFT) filter $\hat{f}_t$ are updated as,
\begin{subequations}
\begin{align}
\hat{g}^k_t &= (1-\gamma)\hat{g}^k_{t-1} + \gamma \overline{\hat{y}_t} . \hat{x}^k_t \\
\hat{h}^k_t &= (1-\gamma)\hat{h}^k_{t-1} + \gamma \Bigg( \sum_{k'=1}^{n_C} \overline{\hat{x}^{k'}_t} . \hat{x}^{k'}_t + \lambda \Bigg)
\end{align}
\label{eq:dcf_model_update}
\end{subequations}
in which the `hat' denotes the 2D DFT, the `bar' denotes complex conjugation, `$.$' denotes pointwise multiplication, $\gamma \in [0, 1]$ is learning rate and $n_C$ is the number of channels.
Next, the filter is constructed by a point-wise division
$\hat{f}^k_t = \hat{h}^k_t / \hat{g}^k_t$.

To locate the target at frame $t$, a sample patch $s_t$ is extracted at the previous location. The filter is applied by computing the correlation scores in the Fourier domain $ \mathscr{F}^{-1} \Big\{ \sum^{n_C}_{k'=1} \overline{\hat{f}^{k'}_{t-1}} . \hat{s}^{k'}_t \Big\}$, in which $\mathscr{F}^{-1}$ denotes the inverse DFT. 

\subsection{Incorporating Information of CNN Layers}
Here, we extend the DCF tracker formulation to accept a linear combination of multiple CNN layers $l \in \mathcal{L}$ with dimensions  $n_W^{[l]} \times n_H^{[l]}\times n_C^{[l]}$. We embed spatial focus, style consistency, temporal coherency, and ST style preserving terms as regularizations over the minimization problem.
\begin{align}
\epsilon = \sum_{\tau=1}^t  \alpha_\tau \left( \sum_{l \in \Lset} a_t^{[l]} ||S (f_t^{[l]}, x_\tau) - y_\tau||^2 + \sum_{x \in \{\mathrm{msk,sty,tmp,sts} \}} \lambda_x R_x(f_t,x_\tau) \right)
\label{eq:ccnt_loss}
\end{align}
where the desired filter form for all layers $l \in \Lset$, $\mathcal{A}_t = \{a^{[l]}_t\}$ is the activation importance of the layers $l$, and $\Lambda = \{ \lambda_{\mathrm{msk}}, \lambda_{\mathrm{sty}}, \lambda_{\mathrm{tmp}}, \lambda_{\mathrm{sts}}\}$ are the regularization weights for tracker components.

\subsection{Regularizing the Filter}
We embed five different regularizations to push the resulting filter toward ideal form given the features of the tracker. To localize the effect of features a mask reg $R_{\mathrm{msk}}$ is used, to penalize the style mismatches between target and the template, co-incidental reg $R_{\mathrm{sty}}$ is employed, to push temporal consistency of the target and smoothness of tracking, the temporal reg $R_{\mathrm{tmp}}$ is proposed, and to punish abrupt ST changes, the ST style reg $R_{\mathrm{sts}}$ is introduced.
\subsubsection{Mask Component}
To address the boundary problems induced by the periodic assumption \cite{danelljan2015learning} and minimizing the effect of background \cite{jepson2003robust} we use mask regularization to penalize filter coefficients located further from object's center:
\begin{equation}
R_{\mathrm{msk}}(f_t, x_\tau) = \sum_{l \in \Lset} b_t^{[l]} \sum_{k=1}^{n_C^{[l]}}||\mathbf{w}.f^{k,[l]}_t||^2
\label{eq:spatial_reg}
\end{equation}
in which $\mathbf{w}: \{1,\ldots,n_W^{[l]}\}\times \{1,\ldots,n_H^{[l]}\} \rightarrow [0,1]$ is the spatial penalty function, and $\mathcal{B}_t = \{b^{[l]}_t\}$ is the spatial importances of the layers. We use Tikhonov reg. similar to \cite{danelljan2015learning} as $\mathbf{w}$ smoothly increase with distance from the target center. 

\subsubsection{Co-incidental Component}
CNN activations entails spatial information of the target but may suffer from extreme target deformations, missing information in the observation (e.g. due to partial occlusions) and complex transformations (e.g. out-of-plane rotations). On the other hand, Gram-based description of a CNN layer encodes the second order statistics of the set of CNN filter responses and tosses spatial arrangements \cite{berger2016incorporating}. Although, this property may lead to some unsatisfying results in NST domain, it is desired in the context of visual tracking as a complement for raw activations.
\begin{equation}
R_{\mathrm{sty}}(f_t,x_\tau) = c_{\mathrm{norm}} \sum_{l \in \Lset} c_t^{[l]} ||G^{[l]}(f_t) - G^{[l]}(f_\tau)||^2_F
\label{eq:style_reg}
\end{equation}
where $\mathcal{C}_t = \{c^{[l]}_t\}$ is the layers' co-incidental importance, $c_{\mathrm{norm}} = \sum_{l \in \Lset} c_t^{[l]} = (2n_H^{[l]}n_W^{[l]}n_C^{[l]})^{-2}$ as normalizing constant and $||.||^2_F$ is the Frobenios norm operator and $G^{[l]}(.)$ is the cross-covariance of activations in layer $l$, the Gram matrix:
\begin{equation}
G^{[l]}_{kk'} = \sum^{n_H^{[l]}}_{i=1} \sum^{n_W^{[l]}}_{j=1} q^{[l]}_{ijk} q^{[l]}_{ijk'}
\label{eq:gram}
\end{equation}
where $q^{[l]}_{ijk}$ is the activation of neuron in $(i,j)$ of channel $k$ in layer $l$. The Gram matrix captures the correlation of activations across different channels of layer $l$, indicating the frequency of co-activation of features of a layer. It is a second-degree statistics of the network activations, that captures co-occurrences between neurons, known as ``style'' in spatial domain. While network activations reconstruct the image based on the features in each layer, style information encodes input patterns, which in lowest form is considered as the texture, known to be useful for tracking \cite{wiyatno2019physical}. The patterns of deeper layers contain higher levels of co-occurrences, such as the relation of the body-parts and shape-color. 

\subsubsection{Temporal Component}
This term is devised to ensure the smoothness of activation alterations of CNNs, which means to see the same features in the same positions of the input images, and punish big alterations in the target appearance, which may happen due to misplaced sampling window. Another benefit of this term is to prefer bounding boxes which include all of the features and therefore improve the scale adaptation ($\mathcal{D}_t = \{d^{[l]}_t\}$): 
\begin{equation}
R_{\mathrm{tmp}}(f_t,x_\tau) = \sum_{l \in \Lset} d_t^{[l]} ||S (f_t^{[l]}, x_\tau) - S (f_t^{[l]}, x_{\tau-1})||^2
\label{eq:temporal_reg}
\end{equation}
\subsubsection{Spatiotemporal Style Component}
To capture the style of target's ST changes, the style of the spatial patterns in consecutive frames is compared. It promotes the motion smoothness of the spatial features, and monitors the style in which each features evolve throughout the video ($\mathcal{E}_t = \{e^{[l]}_t\}$):
\begin{equation}
R_{\mathrm{sts}}(f_t,x_\tau) = \sum_{l \in \Lset} e_t^{[l]} ||G^{[l]}(f_\tau) - G^{[l]}(f_{\tau-1})||^2_F
\label{eq:sts_reg}
\end{equation}
\subsubsection{Model Update}
Extending filter update equations (eq\eqref{eq:dcf_model_update}) to handle multiple layers is not trivial. It is the importance weights $\mathcal{A}_t,\ldots,\mathcal{E}_t,$ that provides high degree of flexibility for the visual tracker to tailor its use of different layers (i.e., their activations, styles, spatial, temporal, and spatitemporal coherences) to the target. As such, we use a simple yet effective way of adjusting the weights, considering the effect of the layer they represent among all the layers. Here, we denote $\tilde{z}_t^{[l]}$ as the portion of error in $t$ caused by layer $l$ among all layers $\Lset$:
\begin{equation}
z^{[l]}_{t+1} = 1 - \frac{\eta + \tilde{z}_t^{[l]}}{\eta + \sum_{l' \in \Lset} \tilde{z}_t^{[l']}} \quad, \mathrm{where} \; z_t \in \{ a_t, b_t, c_t,d_t,e_t\}
\label{eq:parameter_update}
\end{equation}
in which $\eta$ is a small constant and error terms $\tilde{z}_t^{[l]}$ are defined as follows:
\begin{subequations}
\begin{align}
\tilde{a}_t^{[l]} &= ||S (f_t^{[l]}, x_t) - y_t||^2 \\
\tilde{b}_t^{[l]} &= a^{[l]}_t\tilde{a}_t^{[l]} + \sum\nolimits_{k=1}^{n_C^{[l]}}||w.f^{k,[l]}_t||^2 \\
\tilde{c}_t^{[l]} &= a^{[l]}_t\tilde{a}_t^{[l]} + b_t^{[l]}\tilde{b}_t^{[l]} +  ||G^{[l]}(f_t) - G^{[l]}(f_{t-1})||^2_F \\
\tilde{d}_t^{[l]} &= a^{[l]}_t\tilde{a}_t^{[l]} + b_t^{[l]}\tilde{b}_t^{[l]} + c^{[l]}_t\tilde{c}_t^{[l]} + ||S (f_t^{[l]}, x_t) - S (f_t^{[l]}, x_{t-1})||^2\\
\tilde{e}_t^{[l]} &= a^{[l]}_t\tilde{a}_t^{[l]} + b_t^{[l]}\tilde{b}_t^{[l]} + c^{[l]}_t\tilde{c}_t^{[l]} + d^{[l]}_t\tilde{d}_t^{[l]} + ||G^{[l]}(f_t) - G^{[l]}(f_{t-1})||^2_F
\end{align}
\label{eq:parameter_update_loss}
\end{subequations}
In the update phase, first, $\tilde{a}_t$ is calculated that represents the reconstruction error. Plugged into eq\eqref{eq:parameter_update} (which is inspired by AdaBoost), $a_{t+1}$ is obtained. $a_t$ for layer $l$ is the weight of reconstruction error of this layer compared to the other layers, which is weighted by its importance $\tilde{a}_t$. Next, the weighted reconstruction error is added to the raw mask error to give the $\tilde{b}_t$. This is, in turn, used to calculate the weight of the mask error in this layer. This process is repeated for coincidental error, temporal component, and ST component. The errors of each layer are also accumulated to update the weight of the next. Hence, the network won't rely on the style information of a layer with large reconstruction error, etc. The same holds for ST co-occurrences.
\subsubsection{Optimization and Target Localization}
Following \cite{danelljan2015learning}, we used the Gauss-Seidel iterative approach to compute filter coefficients. The cost can be effectively minimized in Fourier domain due to the sparsity of DFT coefficients after regularizations. Image patch with the minimum error of eq\eqref{eq:ccnt_loss} is a target candidate and target scale is estimated by applying the filter at multiple resolutions. The maximum filter response corresponds to the target's location and scale.

\subsection{Implementation Details}
We used VGG19 network consisting of 16 convolutional and 5 max-pooling layers, as implemented in MatConvNet\cite{vedaldi2015matconvnet} and pre-trained on the ImageNet dataset for the image classification. 
To be constistant with \cite{DeepSRDCF} and \cite{gatys2016image}, we used the conv layers after the pooling 
. We also added the input as Layer 0 which enables the tracker to benefit from NCC tracking, hence $\Lset = \{$\texttt{\small input, conv1\_1, conv2\_1, conv3\_1, conv4\_1, conv5\_1}$\}$.
In our implementation, $\sum_{l \in \Lset} a_t^{[l]}, \ldots, e_t^{[l]}=1$, regularization weights $\Lambda$ are determined with cross-validation on YouTubeBB \cite{real2017youtube} and are fixed for all the experiments, others parameters are similar to \cite{DeepSRDCF}.

\section{Experiments}
We take a step-by-step approach to first prove that adding co-incidental and temporal regularization to the baseline improves the tracking performance, and then show that combining multiple layers can improve the tracking performance significantly. We also show that the regularization based on activation, style, and temporal coherence is helpful only if proper importance parameters are selected for different layers. Then we discuss the effect of different regularization terms on the performance of the tracker in different scenarios. Finally, we compare our proposed algorithm with the state-of-the-art and discuss its merits and demerits.
For this comparison, we used success and precision plots and PASCAL metric ($IoU > 0.50$) over OTB50\cite{wu2013online}. For each frame, the area of the estimated box divided by the area of the annotation is defined as \textit{scale ratio}, and its average and standard deviation represents the scale adaptation and jitteriness of a tracker. 

For the comparison with latest trackers, we use OTB100 \cite{wu2015object}, UAV123 \cite{mueller2016benchmark} and LaSot \cite{fan2019lasot} with success and precision indexes and VOT2015 \cite{kristan2015visual} and VOT2018 \cite{kristan2018sixth} using accuracy, robustness, and expected average overlap (EAO).\footnote{More info: \url{http://ishiilab.jp/member/meshgi-k/ccnt.html}.} We have developed our tracker with Matlab using MatConvNet and C++ and on a Nvidia RTX2080 GPU, we achieved the speed of 53.8 fps.

\subsection{The Effects of Regularization on Single Layer} 
In this experiment, we study the effect of proposed regularizations on different CNN layers, used as the features in our baseline tracker, the single layer DeepDCF using eq\eqref{eq:dcf_loss}. Mask regularization (eq\eqref{eq:spatial_reg}) as MR, proposed co-incidental (CR, eq\eqref{eq:style_reg}), temporal (TR, eq\eqref{eq:temporal_reg}) and ST (SR, \eqref{eq:sts_reg}) are then progressively added to the baseline tracker to highlight their contribution in the overall tracker performance (all importance weights are fixed to 1). 

\begin{table}[h]
\caption{The effectiveness of regularizations with single layer of CNN with success rate $IoU > 0.50$. Here, we benchmarked baseline (B) with mask (MR), co-incidental (CR), temporal (TR), and ST style (SR) regularizations on OTB50\cite{wu2013online}. 
}
\label{tab:eval_single_layer}
\centering
\scalebox{0.75}{
\renewcommand{\arraystretch}{1.1}
\begin{tabular}{@{}l c c c c c c@{}}
\hline
Layer ($l$)   	& L0         & L1     & L2     & L3     & L4     & L5     \\ \hline
B             	& 46.2       & 62.3    & 57.4    & 53.9    & 52.9    & 56.3  \\
B + MR          & \textbf{49.5}        & 65.1    & 60.0    & 56.4  & 55.0     & 58.5     \\
B + CR          & 45.1       & 61.9     & 57.5     & 57.4     & 55.1     & 58.4     \\
B + TR          & 45.8       & 62.2    & 57.9     & 55.6     & 54.0     & 57.8    \\
B + MR + CR     & 46.2 		 & 62.7    & 59.0     & 55.9     & 54.8     & 58.3   \\
B + MR+ CR + TR & 47.5       & 64.3    & \textbf{60.1}    & 57.3    & 57.5    & \textbf{62.8} \\
B + MR+ CR + TR + SR & 48.1  & \textbf{64.7}  & \textbf{60.1}  & \textbf{58.3}  & \textbf{58.2}  & 62.0\\
\hline
\end{tabular}
}
\end{table}

\noindent\textbf{Layer-wise Analysis:} Table \ref{tab:eval_single_layer} shows that the activations of features in the shallower layer of CNN generally yields better tracking compared to the deeper layers, except L5 which according to \cite{DeepSRDCF} contains high-level object-specific features. Shallower layers encodes more spatial information while accommodating a certain degree of invariance in target matching process. Contrarily, deeper layers ignore the perturbations of the appearance and perform semantic matching. 

\noindent\textbf{Mask Reg:} Results shows that the use of mask regularization for tracking improve the tracking performance around 2.1-3.3\%, where shallower layers benefit more from such regularization.

\noindent\textbf{Style Reg:} The style information (CR) generally improves the tracking, especially in deeper layers which the activations are not enough to localize the target. However, when applied to shallower layers, especially input image, the style information may be misleading for the tracker which is aligned with the observation of Gatys et al. \cite{gatys2016image}.

\noindent\textbf{Temporal Reg:} Deeper layers enjoys temporal regularization more. This is due to the fact that changes in activations in deeper layers signals semantic changes in the appearance, such as misalignment of the window to the target, partial or full occlusions or unaccounted target transformations such as out-of-plane rotations. In contrary, the changes in shallower layers come from minor changes in the low-level features (e.g. edges) that is abundant in the real-world scenarios, and using only this regularization for shallow layers is not recommended.

\noindent\textbf{Spatiotemporal Reg:} Using this regularization on top of temporal regularization, often improves the accuracy of the tracking since non-linear temporal styles of the features cannot be always handled using temporal reg.

\noindent\textbf{All Regularizations:} The combination of MR and CR terms, especially helps the tracking using deeper layers and starting from L2 it outperforms both MR and CR regularizations. The combination of all regularization terms proved to be useful for all layers, improving tracking accuracy by 2-6\% compared to baseline.

\noindent\textbf{Feature Interconnections}: Feature interconnections can be divided into \textit{(i)} spatial-coincidental (when two features have high filter responses at the same time), \textit{(ii)} spatial-temporal (a feature activates in a frame), \textit{(iii)} spatial-ST style (a feature coactivates with a motion feature), \textit{(iv)} style-temporal (coupled features turns on/off simultaneously), \textit{(v)} style-ST style (coupled features moves similarly), temporal-ST style (a features starts/stops moving). The features are designed to capture different aspects of the object's appearance and motion; they are sometimes overlapping. Such redundancy improves tracking, with more complex features improving semantics of tracking, and low-level features improving the accuracy of the localization.

\subsection{Employing Multiple Layers of CNN}
To investigate different components of the proposed tracker, we prepared several ablated versions and compared them in Table \ref{tab:eval_multi_layer}. 
Three settings have been considered for the importance weights: \textit{uniform weights}, \textit{random weights}, and \textit{optimized weights based on the model update} (eq\eqref{eq:parameter_update}). The random initial weights were generated for each time $t$ (summed up to 1), and the experiment was repeated five times and averaged. By adding each regularization term, the speed of the tracker degrades, therefore, we added the ratio of the custom tracker speed to the baseline (first row) in the last column. It should be noted that when the spatial coefficient $b_t^{[l]}$ are zero, the L2 norm of all filter responses (of all layers) is used to regularize. \textit{Uniform weighting} keeps reg. weights fixed and equal during tracking, \textit{random weighting} assigns random weights to different components of each layer and our \textit{proposed} AdaBoost-inspired weighting penalizes components proportional to their contribution in the previous mistakes.

\begin{table}[h]
\caption{The effect of using multiple CNN layers with various importance weight strategies. This is based on the success rate ($IoU > 0.50$) on OTB50. Last column presents the speed of the ablated trackers (+ model update) compared to baseline (\%).}
\label{tab:eval_multi_layer}
\centering
\scalebox{0.75}{
\renewcommand{\arraystretch}{1.1}
\begin{tabular}{@{}l @{}c c c c@{}}
\hline
Model Update                                            & uniform & random & proposed               & speed (\%) \\ \hline
B ($\mathcal{B}_t = \mathcal{C}_t =\mathcal{D}_t=\mathcal{E}_t = 0$)     &  66.8   & 64.4   &  \textbf{79.2}        & 100.0       \\
B + MR ($\mathcal{C}_t =\mathcal{D}_t=\mathcal{E}_t = 0$)             &  67.3   & 66.4   &  \textbf{81.7}        & 95.2        \\
B + CR ($\mathcal{B}_t =\mathcal{D}_t=\mathcal{E}_t = 0$)                &  69.1   & 69.9   &  \textbf{82.8}        & 83.1        \\
B + TR ($\mathcal{B}_t =\mathcal{C}_t=\mathcal{E}_t = 0$)                &  67.3   & 67.0   &  \textbf{81.1}        & 98.4        \\
B + MR + CR ($\mathcal{D}_t=\mathcal{E}_t = 0$)                         &  68.3   & 72.6   &  \textbf{85.9}        & 80.8        \\
B + MR + CR + TR ($\mathcal{E}_t = 0$)                                  & 69.0   & 73.0   & \textbf{86.5}        & 78.0      \\
B + MR + CR + TR + SR													& 69.2   & 73.3   &  \textbf{86.9}        & 78.7\\
\hline
\end{tabular}
}
\end{table}

\noindent\textbf{Comparing Model Update Schemes:} Table \ref{tab:eval_multi_layer} shows that with the use of proposed model update, different components of the tracker may collaborate to improve the overall performance of the tracker when combining different layers. However, uniform weights for all parameters (equal to $|\Lset|^{-1}$) cannot provide much improvement compared to the single layer DeepDCF, especially when compared to the L1 column of Table \ref{tab:eval_single_layer}. Interestingly, random weights outperform uniform weights when applied to style regularization, which shows that not all layers contain equivalently useful co-incidental information.

\noindent\textbf{Multiple Layers:} By comparing each row of the Table \ref{tab:eval_multi_layer} with the corresponding row of Table \ref{tab:eval_single_layer}, the effect of combining different layers is evident. Comparing the first rows shows the advantage of combining layers without using any regularization. Uniform weights for the layers raise the performance only by 4.5\% (all layers vs. L1), whereas the proposed fusion can boost the combination performance up to 16.9\%. This is a recurring pattern for other rows that show the benefit of the layer combination for activations, as well as regularization terms. 

While our method can be seen as a feature selection/weight tuning, it is crucial to see the tuning procedure as a layer-wise adaptation. In each layer, the effect of different regularization term is determined by its contribution in the loss term. This calculation is isolated from other layers. Features of each layer compete with each other to better represent the target, but cooperate with each other to deliver the best overall representation that can be obtained from that particular layer. Additionally, to use different types of features, we utilize the combination of different layers to balance the detail-semantic trade-off in different tracking scenarios; therefore layers' importance should be adaptively adjusted.

\noindent\textbf{Applying Different Reg:} Similar to the case of single layers, regularization multiple layers is also effective. In case of uniform weights, using CR outperforms MR+CR which indicates that without proper weight adjustment, different regularization cannot be effectively stacked in the tracker. Therefore, it is expected that the proposed adaptive weight can handle this case, as table shows. 

\begin{table}[t]
\caption{Scale adaptation obtained by proposed regularizations on OTB-100 measured by the mean of estimate-to-real scale ratio and its standard deviation (jitter).}
\label{tab:eval_scale}
\centering
\scalebox{0.75}{
\renewcommand{\arraystretch}{1.1}
\begin{tabular}{@{}l  c  c  c  c c  c  c@{}}
\hline
Tracker        & B & B+MR & B+CR & B+TR & B+MR+TR & B+MR+CR+TR &  ALL  \\
\hline
Avg.Ratio     & 92.2 &    93.1 &    93.3 &    93.8 &    93.3 & 94.2&   \textbf{94.7}  \\
Jitter         & 8.17 &    7.13 &    5.81 &     2.66 &    5.11 & 2.40&   \textbf{2.35} \\
\hline
\end{tabular}
}
\end{table}

\subsection{Scale Adaptation} 
The proposed style and temporal regs, tend to discard candidates with mismatching scale due to style and continuity inconsistencies. Additionally, temporal reg tend to reduce the jittering of the position and scale. Table \ref{tab:eval_scale} demonstrates the proposed tracker with multi-layers of CNN, adaptive weights and different regs. 

\begin{figure}[t]
\centering
\includegraphics[width=0.45\linewidth]{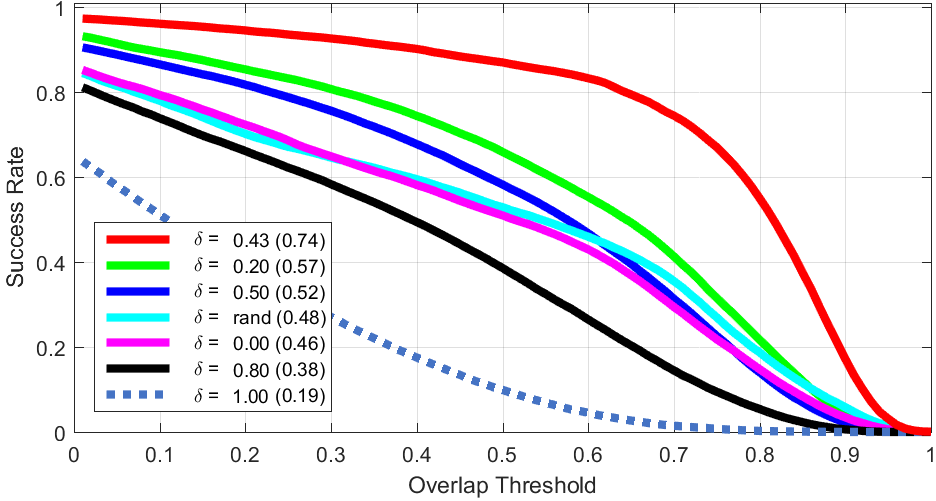}
\includegraphics[width=0.45\linewidth]{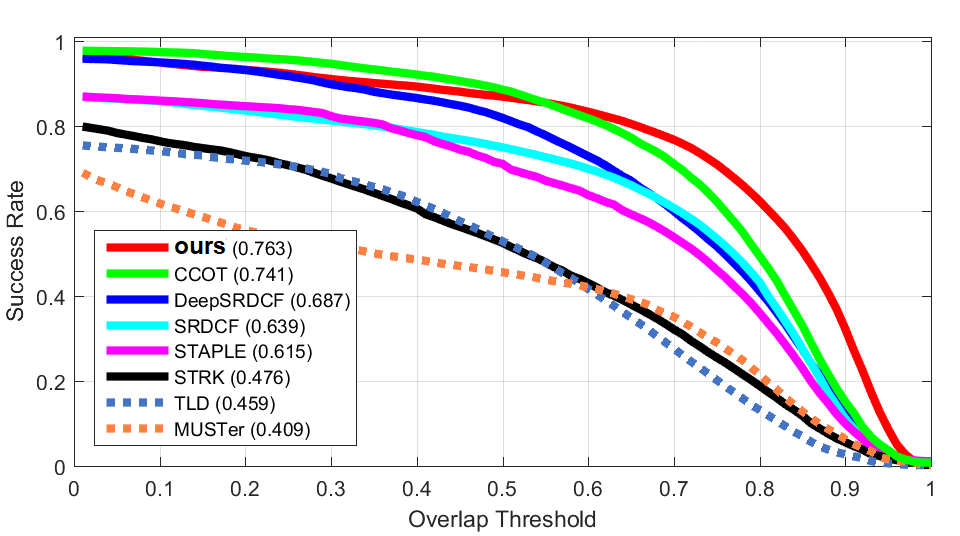}
\caption{(left) The activation vs. style trade-off for the custom tracker on OTB50. While $\delta \rightarrow 1$ puts too much emphasize on the style, $\delta = 0$ overemphasizes on the activations. (middle) Performance comparison of trackers on OTB100 using success plot.}
  \label{fig:eval_style}
\end{figure}
\subsection{Activation vs. Style}
As seen in NST literature \cite{gatys2016image,johnson2016perceptual}, various amount of focus on the content image and style image yields different outcomes. In tracking, however, the accuracy provides a measure to balance this focus. We conducted an experiment to see the effect of the regularization weights $\lambda_{\mathrm{sty}}$ on the tracking performance. Hence, we set $\lambda_{\mathrm{sty}} = \delta$ while disabling spatial and temporal regularizers.  Figure \ref{fig:eval_style}-left depicts the success plot for several $\delta$ and the optimal value $\delta^*$ (via annealing and cross-validation on OTB-50, with proposed model update for layers in $\mathcal{L}$).

This figure also depicts the performance of the obtained tracker with various values of $\delta$. The results reveal that when the tracker ignores the style information ($\delta = 0$, the base multi-layer tracker) its performance is better than when it ignores activations ($\delta = 1$) since the style information is not suitable in isolation for tracking. The values between these two extremes work better by enhancing the activations with style information. However, finding a sweetspot is difficult and scenario-dependent, e.g., for textureless object \cite{choi20123d} more spatial and semantic information is required, whereas textured objects benefit from style feedback. 

\subsection{Preliminary Analysis}
We compared our tracker with TLD \cite{kalal2012tracking}, STRUKK \cite{hare2011struck}, MEEM \cite{zhang2014meem}, MUSTer \cite{hong2015multi}, STPL \cite{bertinetto2016staple}, CMT \cite{meshgi2017active}, SRDCF \cite{danelljan2015learning}, dSRDCF\cite{danelljan2015convolutional} and CCOT \cite{CCOT}.

\begin{table}[!h]
\caption{Quantitative evaluation of trackers (top) using average success on OTB50 \cite{wu2013online} for different tracking challenges; (middle) success and precision rates on OTB100 \cite{wu2015object}, estimated-on-real scale ratio and jitter; (bottom) robustness and accuracy on VOT2015 \cite{kristan2015visual}. The {\color{red}first}, {\color{green}second} and {\color{blue}third} best methods are shown in color.}
\label{tab:eval_pre}
\centering
\scalebox{0.75}{
\renewcommand{\arraystretch}{1.1}
\begin{tabular}{@{}l l c c c c c c c c c@{}}
\hline
&& {\small TLD} & {\small STRUCK} & {\small MEEM}& {\small MUSTer} & {\small STAPLE} & {\small SRDCF}& {\small dSRDCF}&  {\small CCOT}& {\small Ours} \\ 
\hline \hline
\parbox[t]{4mm}{\multirow{12}{*}{\rotatebox[origin=c]{90}{\small OTB50}}}
&{\small Illumination}		& 0.48 & 0.53 & 0.62 & {\color{blue}0.73} & 0.68 & 0.70 & 0.71 &{\color{green}0.75} & {\color{red}0.80} \\
&{\small Deformation}   	& 0.38 & 0.51 & 0.62 & {\color{blue}0.69} & {\color{green}0.70} & 0.67 &0.67& {\color{blue}0.69} & {\color{red}0.78} \\
&{\small Occlusion}    		& 0.46 & 0.50 & 0.61 & 0.69 & 0.69 & 0.70 & {\color{blue}0.71} &{\color{green}0.76} & {\color{red}0.79} \\
&{\small Scale Changes}		& 0.49 & 0.51 & 0.58 & 0.71 & 0.68 & 0.71 &{\color{blue}0.75}& {\color{green}0.76} & {\color{red}0.82} \\
&{\small In-plane Rot.}  	& 0.50 & 0.54 & 0.58 & 0.69 & 0.69 & 0.70 &{\color{green}0.73}& {\color{blue}0.72} & {\color{red}0.80} \\
&{\small Out-of-plane Rot.} & 0.48 & 0.53 & 0.62 & {\color{blue}0.70} & 0.67 & 0.69 &{\color{blue}0.70}& {\color{green}0.74} & {\color{red}0.81} \\
&{\small Out-of-view}	    & 0.54 & 0.52 & 0.68 & {\color{blue}0.73} & 0.62 & 0.66 &0.66& {\color{green}0.79} & {\color{red}0.81} \\
&{\small Low Resolution}	& 0.36 & 0.33 & 0.43 & 0.50 & 0.47 & 0.58 &{\color{blue}0.61}& {\color{green}0.70} & {\color{red}0.74} \\
&{\small Background Clutter}& 0.39 & 0.52 & 0.67 & {\color{green}0.72} & 0.67 & 0.70 &{\color{blue}0.71}& 0.70 & {\color{red}0.78} \\
&{\small Fast Motion}     	& 0.45 & 0.52 & 0.65 & 0.65 & 0.56 & 0.63 &{\color{blue}0.67}& {\color{green}0.72} & {\color{red}0.78} \\
&{\small Motion Blur}     	& 0.41 & 0.47 & 0.63 & 0.65 & 0.61 & 0.69 &{\color{blue}0.70}& {\color{green}0.72} & {\color{red}0.78} \\
\cline{2-11}
&Average Success    & 0.49 & 0.55 & 0.62 & {\color{blue}0.72} & 0.69 & 0.70 & 0.71& {\color{green}0.75} & {\color{red}0.80} \\
\hline \hline
\parbox[t]{4mm}{\multirow{5}{*}{\rotatebox[origin=c]{90}{\small OTB100}}}
&Average Success    & 0.46 & 0.48 & 0.65 &0.57& 0.62 & 0.64 & {\color{blue}0.69} &{\color{green}0.74} & {\color{red}0.76} \\
&Average Precision      & 0.58 & 0.59 & 0.62 &0.74& 0.73 & 0.71 & {\color{blue}0.81} &{\color{red}0.85} & {\color{red}0.85} \\
&$IoU>0.5$    & 0.52 & 0.52 & 0.62 &0.65& 0.71 & 0.75 & {\color{blue}0.78} &{\color{red}0.88} & {\color{green}0.86} \\
\cline{2-11}
&Average Scale   & 116.4 & 134.7 & 112.1 &-& 110.8 & 88.5 & {\color{red}101.8} &{\color{green}94.0} & {\color{blue}93.7} \\
&Jitter       & 8.2 & 8.7 & 8.2 &-& 5.9 & {\color{blue}4.1} & 4.9 &{\color{red}3.8} & {\color{red}2.3} \\
\hline \hline
\parbox[t]{4mm}{\multirow{2}{*}{\rotatebox[origin=c]{90}{\small VOT}}}
&Accuracy     &-& 0.47  & 0.50 & 0.52 & 0.53 & {\color{green}0.56} & 0.53 &{\color{blue}0.54} & {\color{red}0.76} \\
&Robustness   &-& 1.26  & 1.85 & 2.00 & 1.35 & 1.24 & {\color{blue}1.05} & {\color{green}0.82} & {\color{red}0.65} \\
\hline
\end{tabular}
}
\end{table}

\begin{figure}[h]
\begin{center}
\includegraphics[width= 0.19\linewidth]{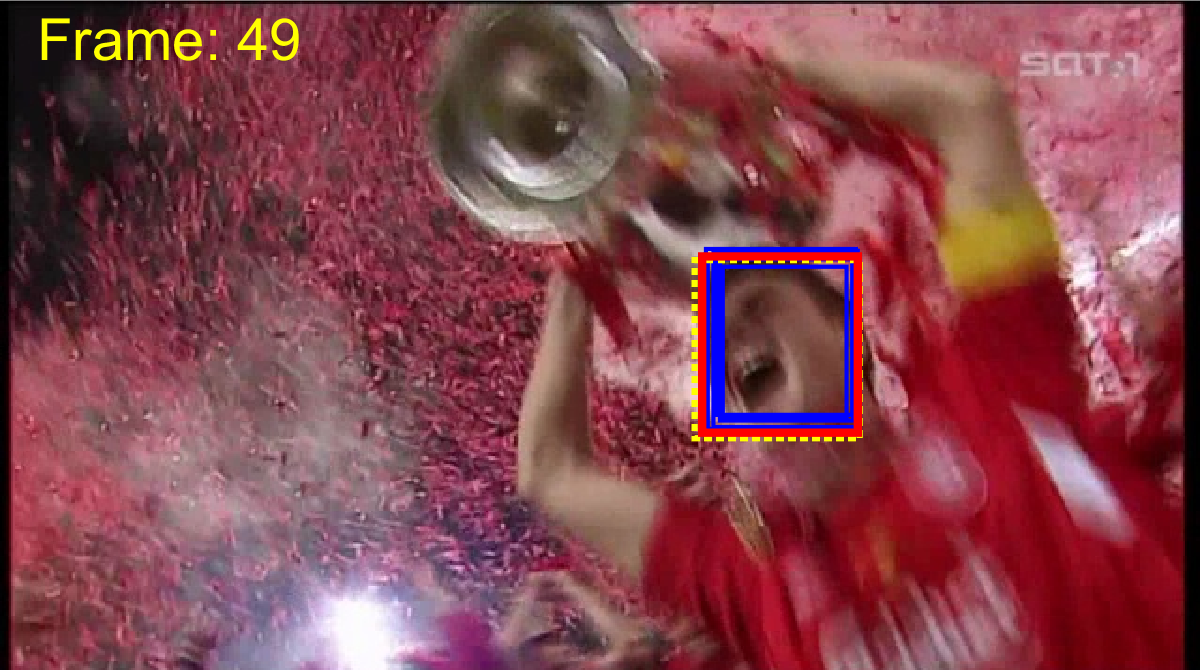}
\includegraphics[width= 0.19\linewidth]{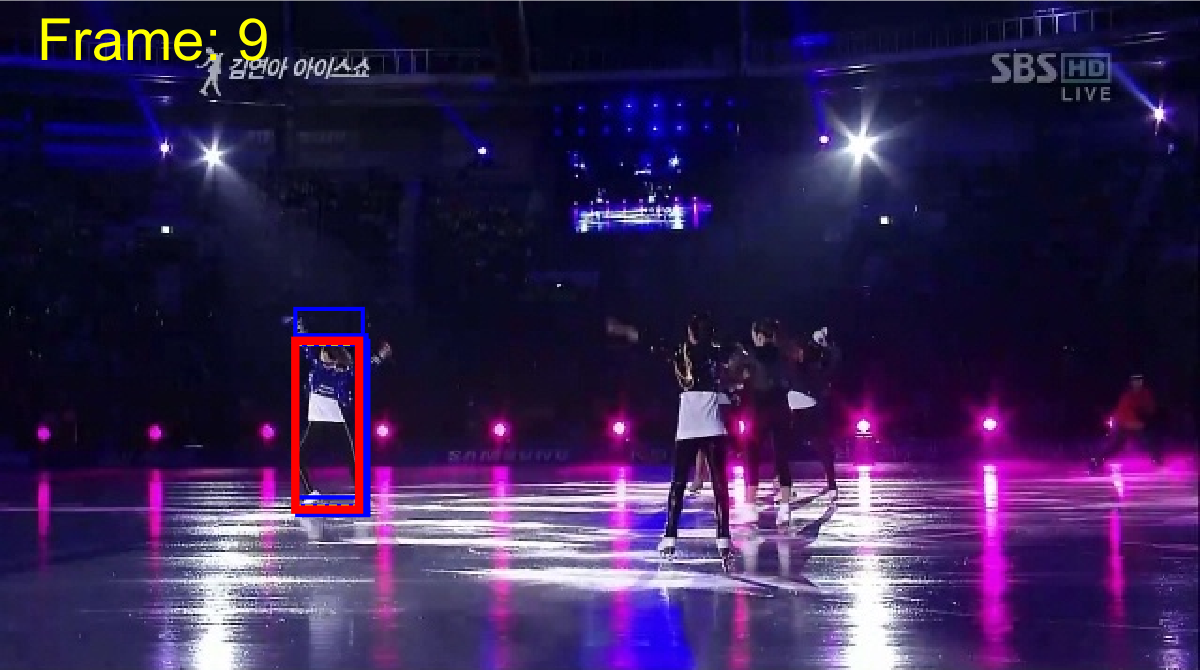}
\includegraphics[width= 0.19\linewidth]{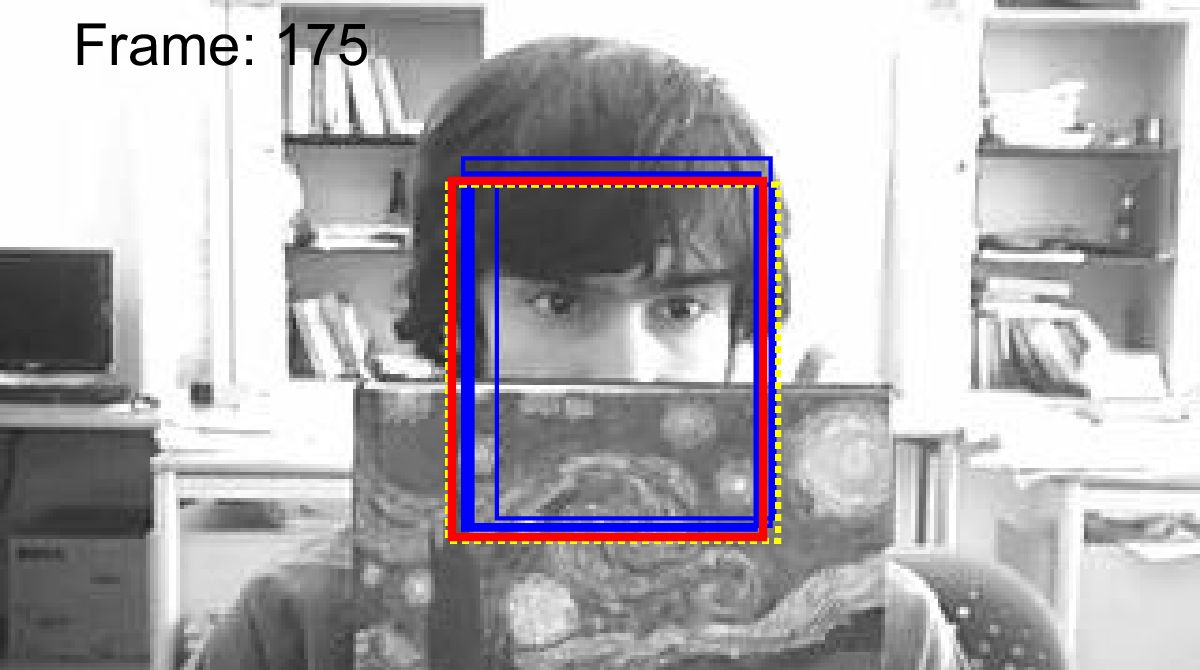}
\includegraphics[width= 0.19\linewidth]{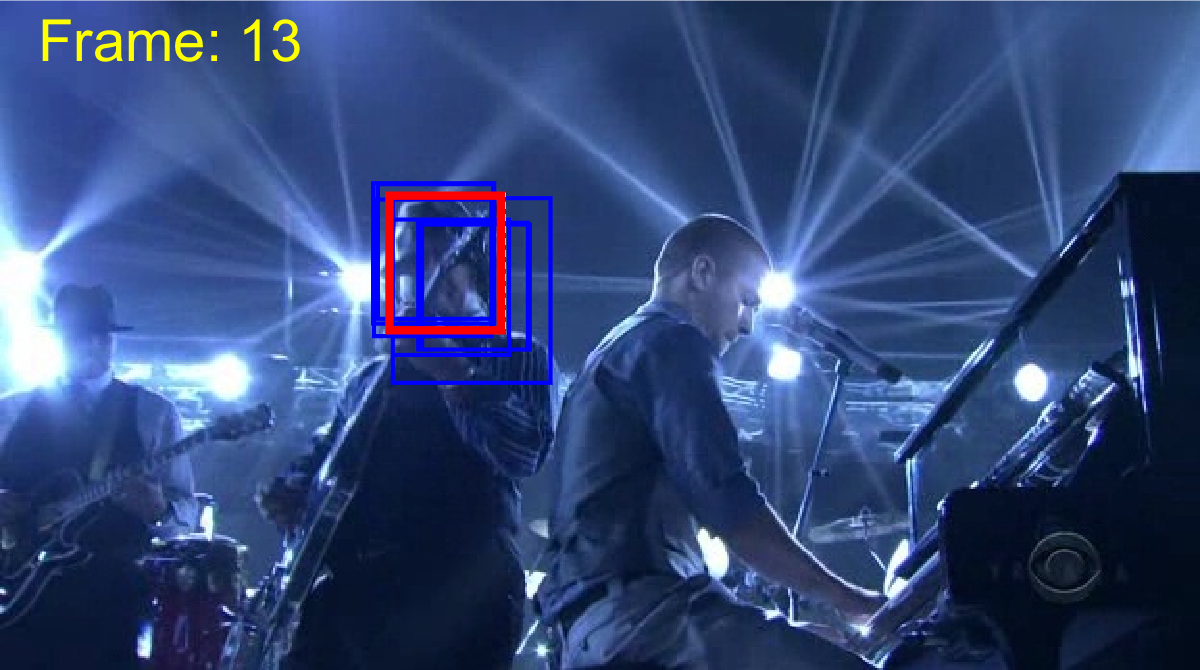}
\includegraphics[width= 0.19\linewidth]{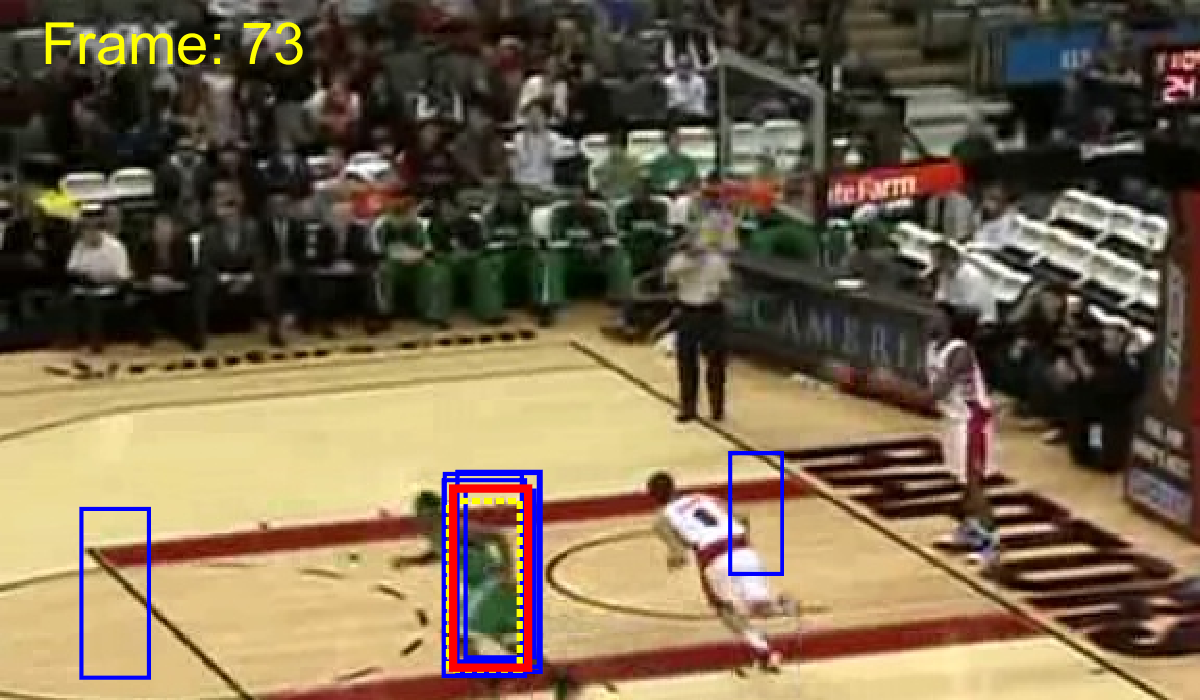}
\\
\includegraphics[width= 0.19\linewidth]{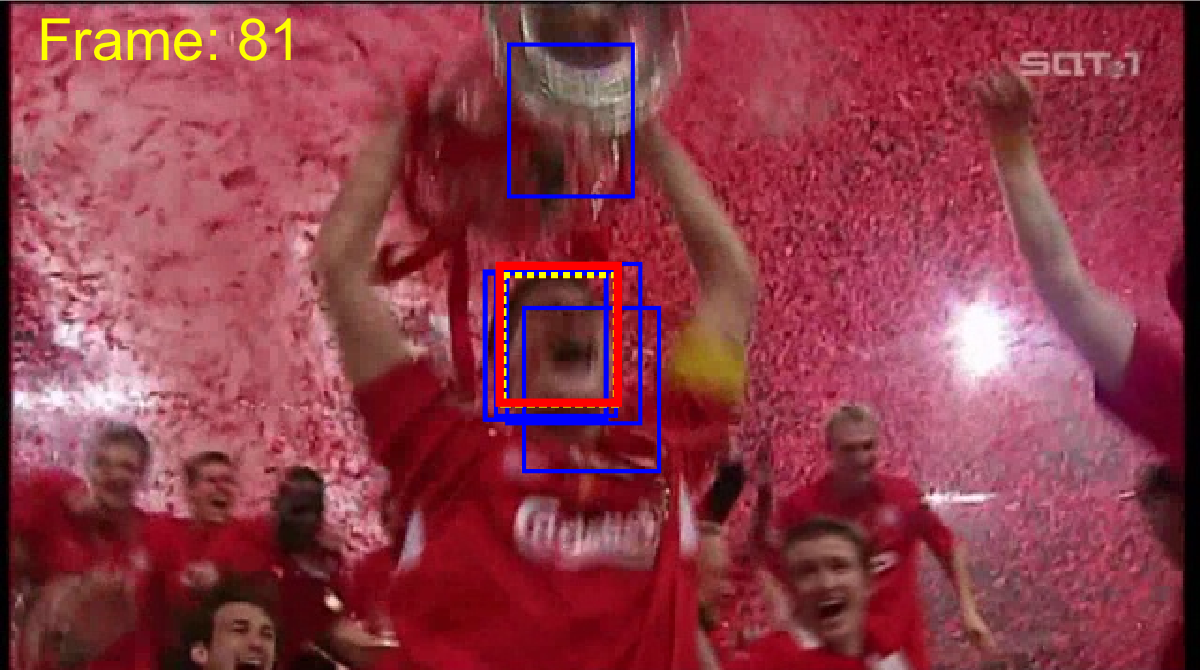}
\includegraphics[width= 0.19\linewidth]{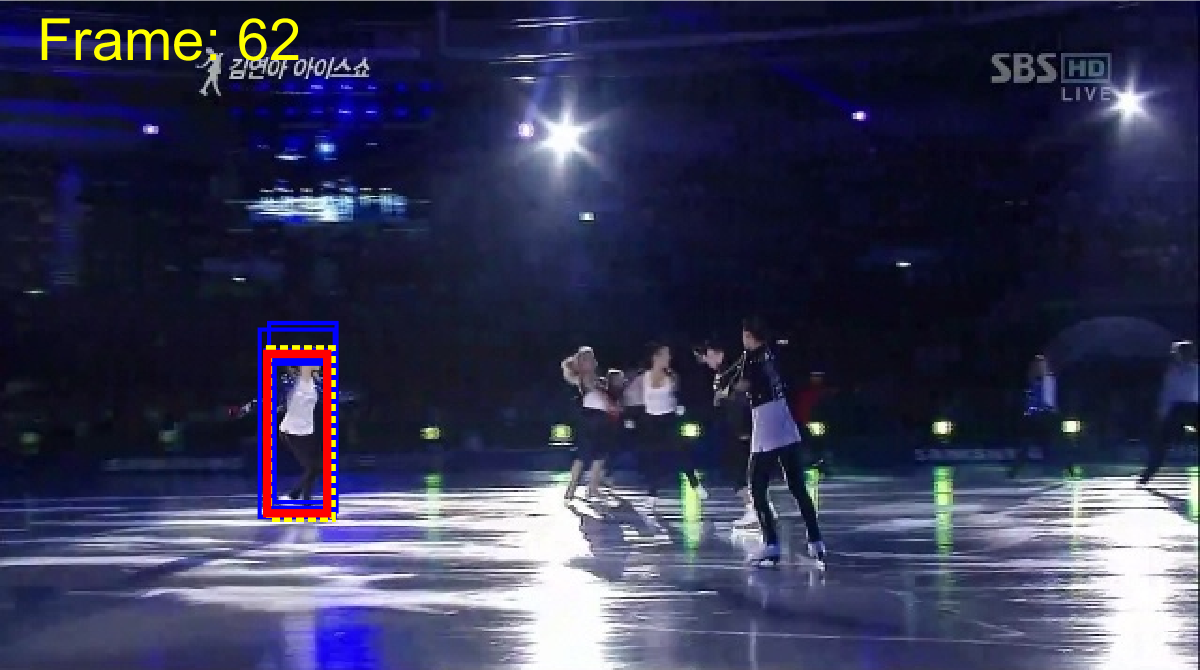}
\includegraphics[width= 0.19\linewidth]{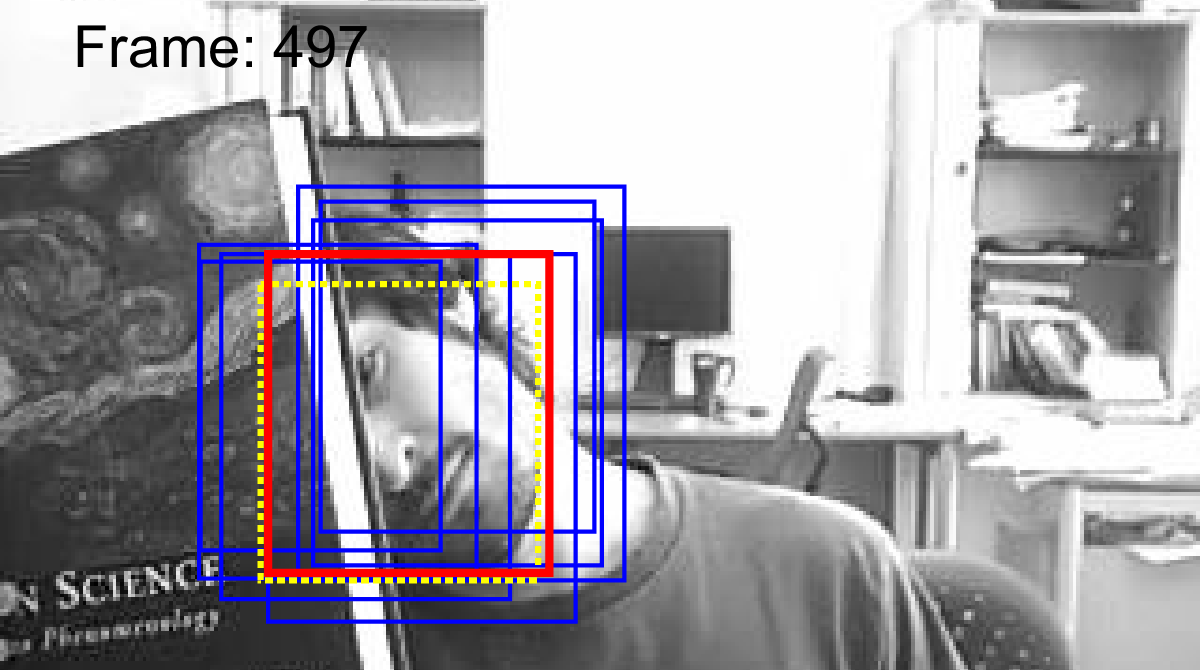}
\includegraphics[width= 0.19\linewidth]{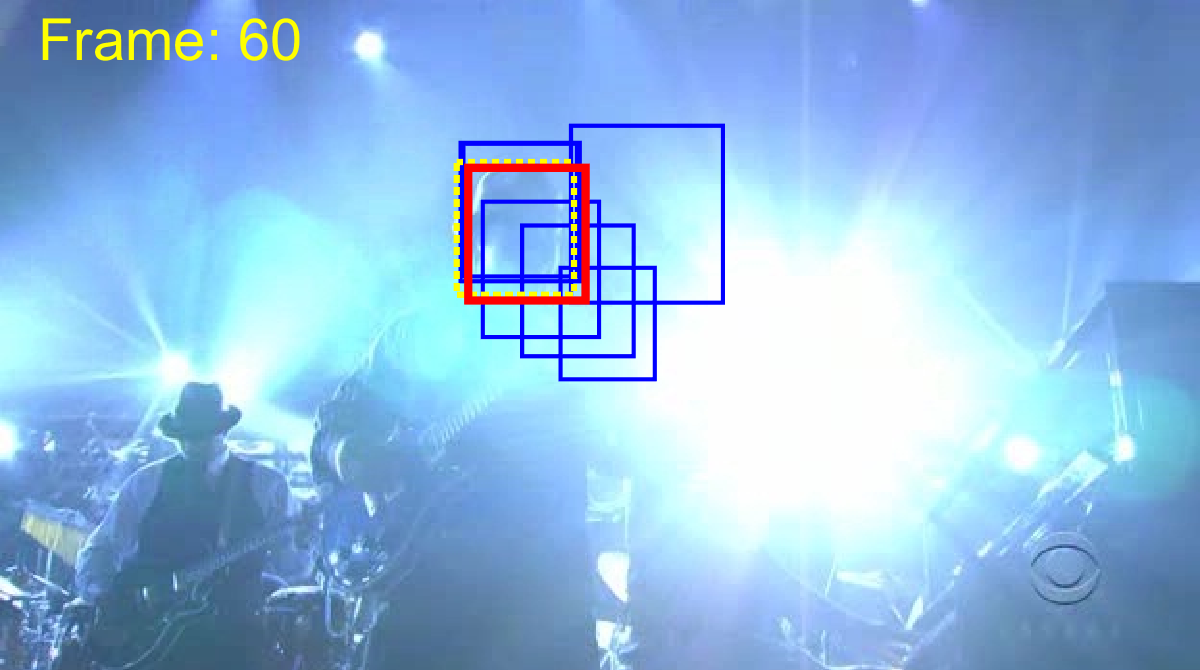}
\includegraphics[width= 0.19\linewidth]{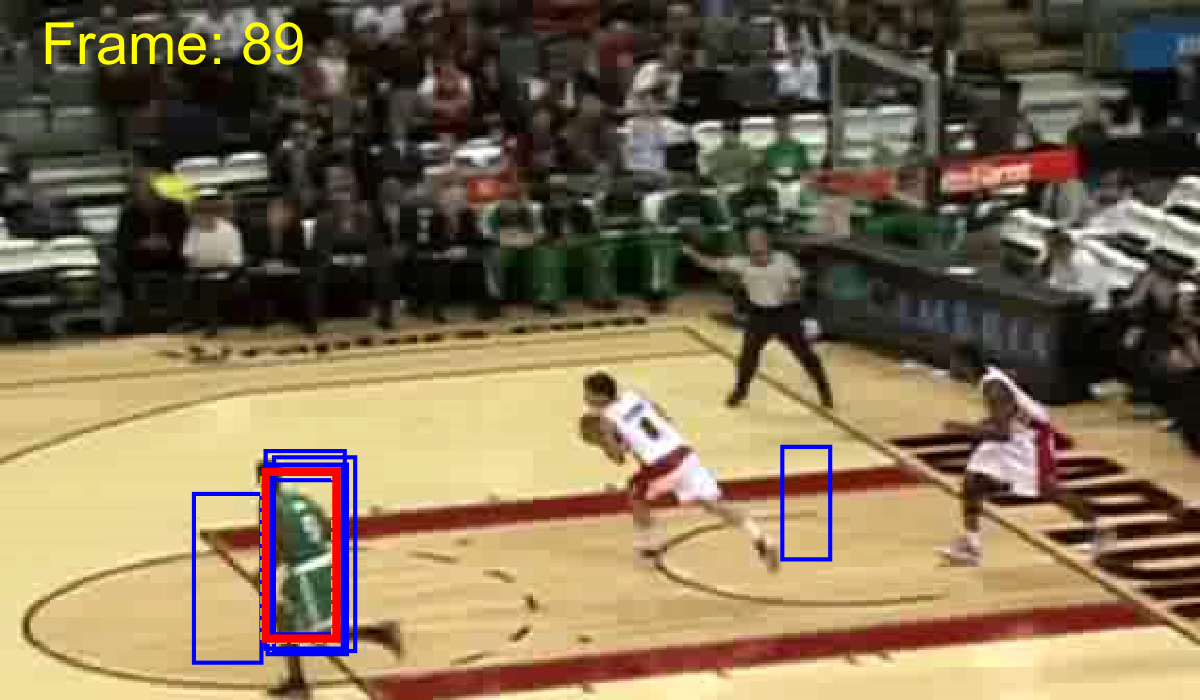}
\\
\includegraphics[width= 0.19\linewidth]{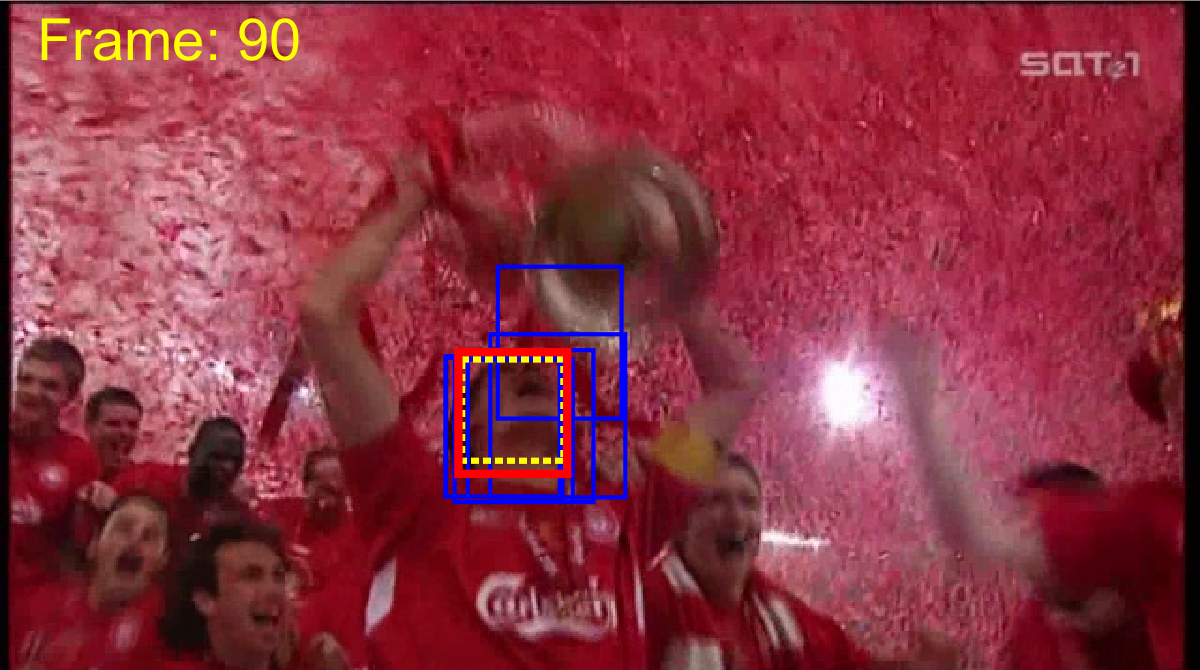}
\includegraphics[width= 0.19\linewidth]{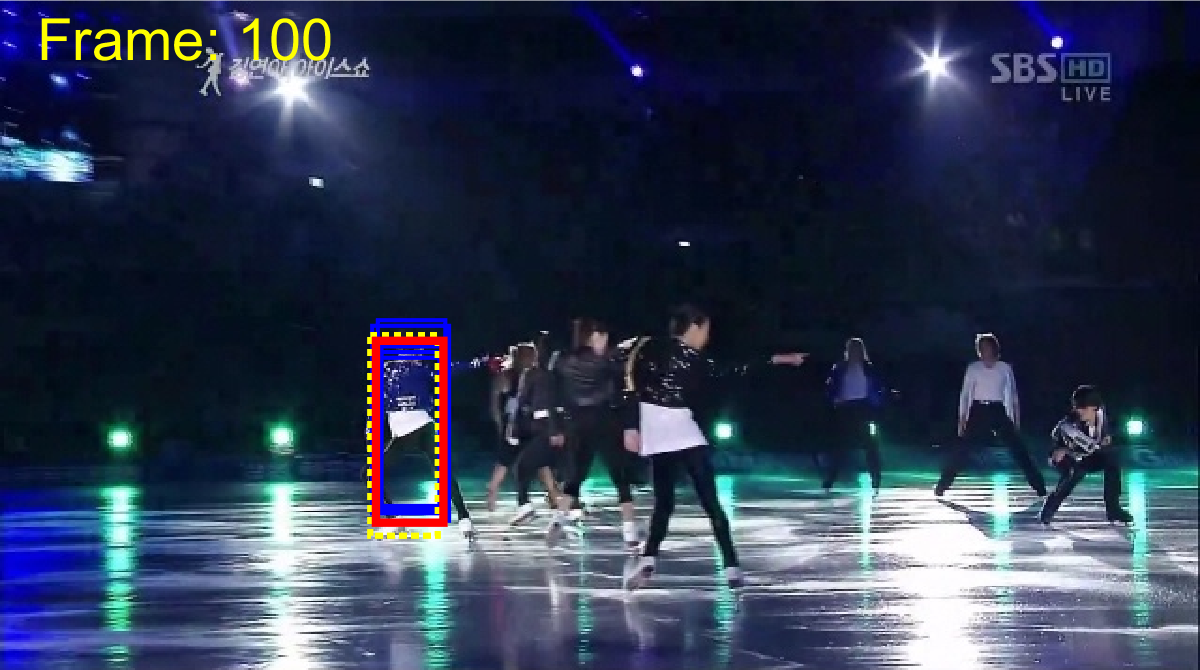}
\includegraphics[width= 0.19\linewidth]{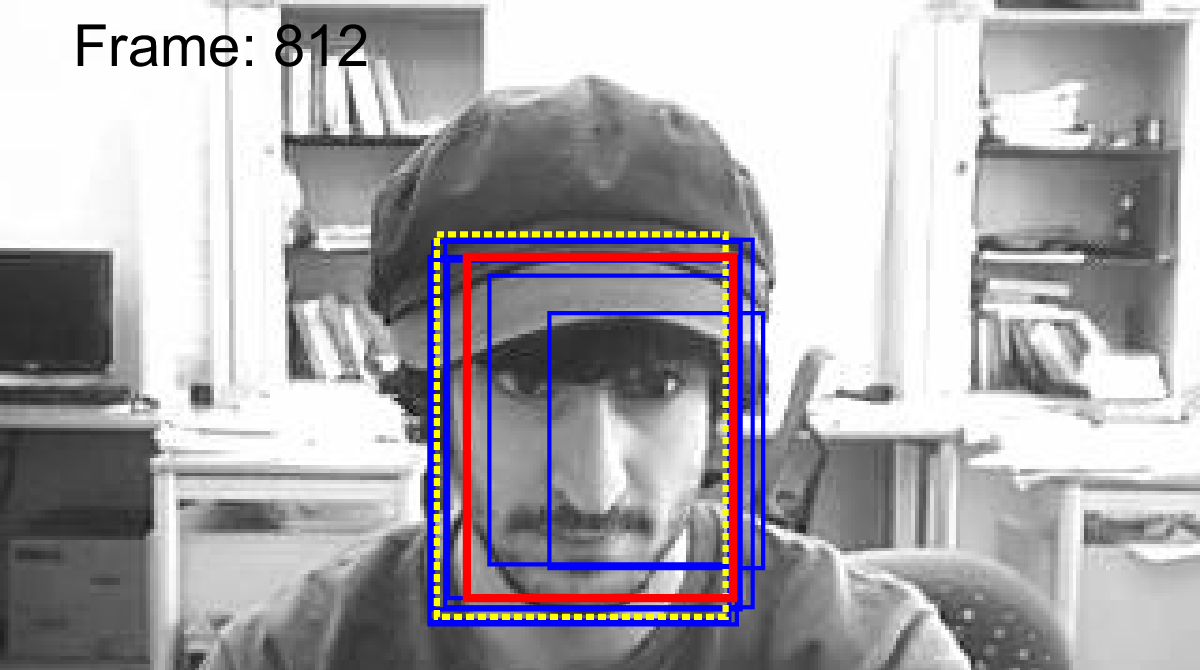}
\includegraphics[width= 0.19\linewidth]{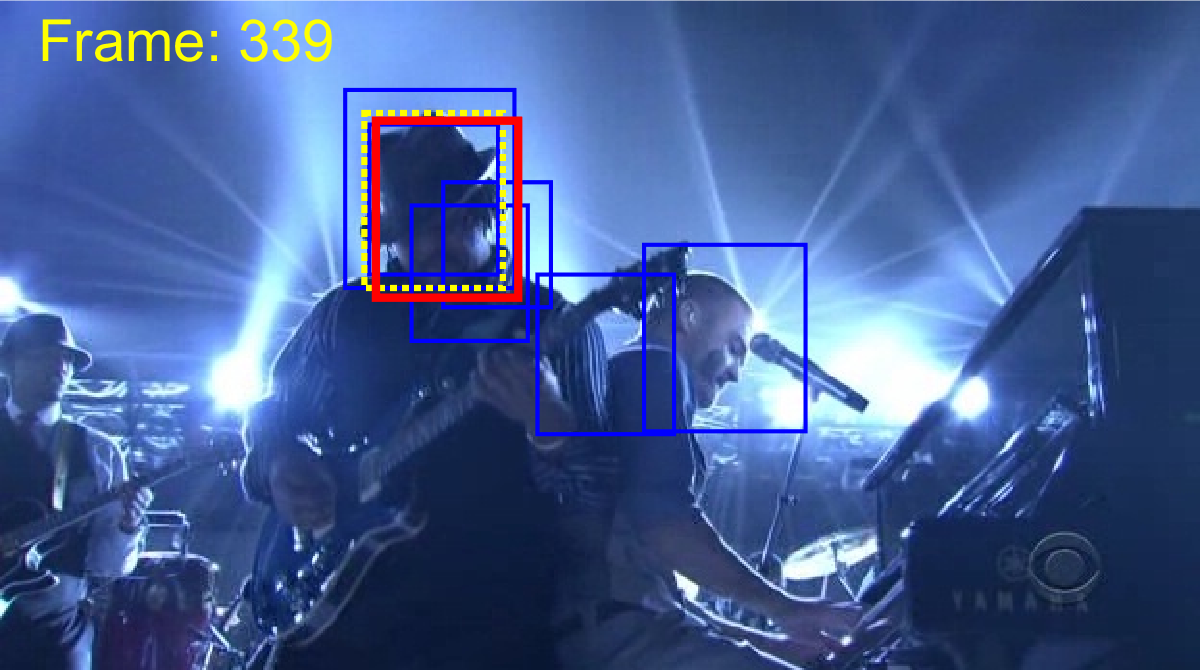}
\includegraphics[width= 0.19\linewidth]{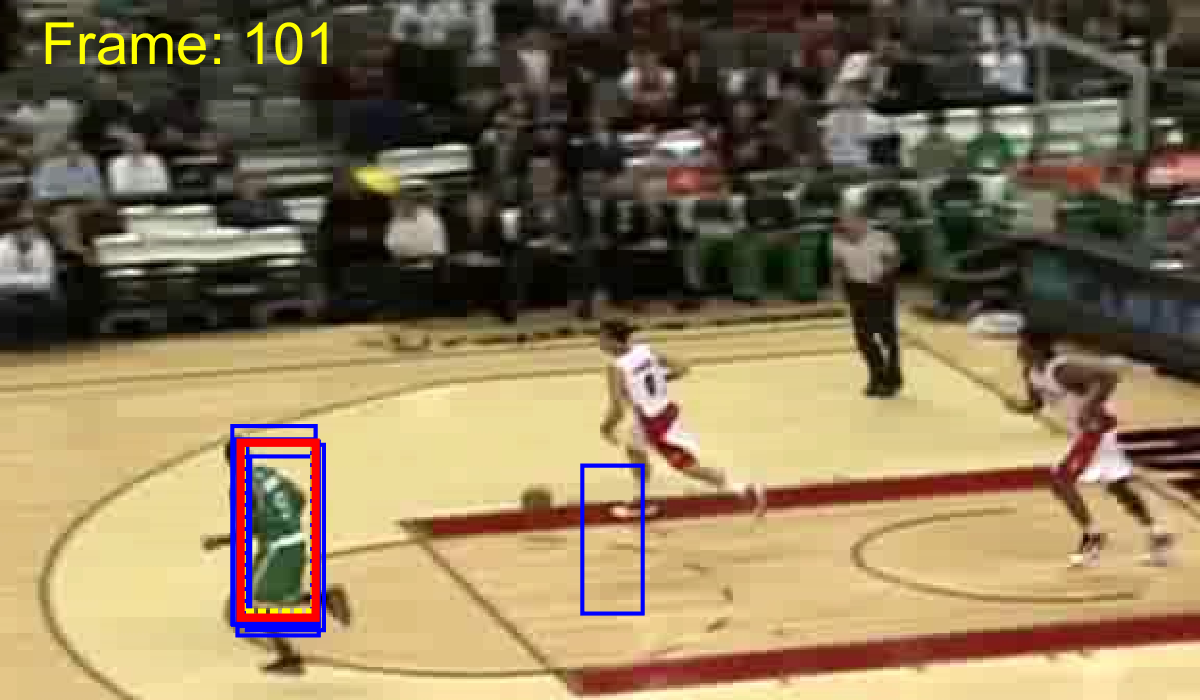}
\end{center}
\caption{Example tracking results on \texttt{Soccer},\texttt{Skating1}, \texttt{FaceOcc1}, \texttt{Shaking}, and \texttt{Basketball} with severe occlusion, noise and illumination changes, scaling and 3D rotations, and clutter. (Red:  proposed tracker, Blue: other trackers, Yellow: GT). 
}
\label{fig:qualitative}
\end{figure}

\noindent\textbf{Attribute Analysis:} 
We use partial subsets of OTB50 \cite{wu2013online} with a distinguishing attribute to evaluate the tracker performance under different situations. Table \ref{tab:eval_pre} shows the superior performance of the algorithm, especially in handling deformations (by adaptive fusion of deep and shallow layers of CNN) and background clutter (by spatial and style reg.) and motion (by temporal reg.). Figure \ref{fig:qualitative} demonstrates the performance of the tracker on several challenging scenarios.

\noindent\textbf{OTB100:} 
Figure \ref{fig:eval_style} (right) and Table \ref{tab:eval_pre} presents the success and precision plots of our tracker along with others. Data shows that proposed algorithm has superior performance, less jitter, and comparable localization and scale adaptation.

\noindent\textbf{VOT2015:} 
Table \ref{tab:eval_pre} also shows superior performance in terms of accuracy coupled with decent robustness.

\subsection{Comparison with State-of-the-Art}
\noindent\textbf{Deep Trackers:} We compared our tracker against recent deep trackers on OTB100, including ECO \cite{danelljan2017eco}, ATOM \cite{danelljan2019atom}, VITAL \cite{VITAL}, HDT \cite{HDT}, YCNN \cite{YCNN}, MDNet \cite{MDNet}, dSTRCF \cite{li2018learning}, STResCF \cite{zhu2019stresnet_cf}, CBCW \cite{zhou2018efficient}, SiamFC \cite{SIAMESEFC}, SiamRPN \cite{li2018high}, SiamRPN++ \cite{li2019siamrpn++}, SINT++ \cite{SINTpp}, and DiMP \cite{bhat2019learning}. Table \ref{tab:eval_deep_otb50} shows that although our proposed tracker has some issues in accurate localization, it has a superior overall performance and success rate in handling various scenarios.
%

The experiments revealed that the proposed tracker is resistant to target abrupt or extensive target changes. Temporal and ST features in our method monitor the inconsistency in target appearance and style, co-occurrence features in different levels of abstraction provide different levels of robustness to target changes (from low-frequency features to the high-level features such as object part relations. The dynamic weighting enables the tracker to have the flexibility to resort to more abstract feature when the target undergoes drastic changes, and ST features handle abnormalities such as temporal occlusion and deformations. 
%

\begin{table}[t]
\caption{Evaluation of deep trackers on OTB100 \cite{wu2015object} using success rate and precision.} 
\label{tab:eval_deep_otb50}
\centering
\scalebox{0.7}{
\renewcommand{\arraystretch}{1.1}
\begin{tabular}{@{}l ccccccccccccccccc@{}}
\hline
 & {\small ECO}& {\small ATOM} & {\small VITAL}& {\small HDT} & {\small YCNN}& {\small MDNet}& {\small dSTRCF}& {\small STResCF}& {\small CBCW}& {\small SiFC} & {\small SiRPN}& {\small SiRPN++}& {\small SINT++}& {\small \textbf{Ours}}\\ \hline
Avg. Succ    &{\color{green}0.69}&0.66&0.68&0.65&0.60&0.67&0.68&0.59&0.61&0.59&0.63&{\color{green}0.69}&0.57&{\color{red}0.76}\\
Avg. Prec    &{\color{red}0.91}& - &{\color{red}0.91}&0.84&0.84&{\color{green}0.90}& -    &0.83&0.81&0.78&{\color{blue}0.85}&{\color{red}0.91}&0.76&{\color{blue}0.85}\\
$IoU>\frac{1}{2}$    &0.74&{\color{red}0.86}&0.79&0.68&0.74&{\color{green}0.85}&0.77&0.76&0.76&0.76&0.80&{\color{blue}0.83}&0.78&{\color{red}0.86}\\
\hline
\end{tabular}
}
\end{table}

\begin{table}[h]
\caption{Evaluation on VOT2018 by the means of EAO, robustness and accuracy.}
\label{tab:eval_vot18}
\centering
\scalebox{0.7}{
\renewcommand{\arraystretch}{1.1}
\begin{tabular}{@{}lccccccccccc@{}}
\hline
& {\small STURCK} & {\small MEEM}& {\small STAPLE} & {\small SRDCF} & {\small CCOT} &{\small SiamFC}&  {\small ECO}& {\small SiamRPN}& {\small SiamRPN++}& {\small ATOM}&{\small \textbf{Ours}} \\ \hline
EAO         & 0.097 & 0.192 & 0.169 & 0.119 & 0.267 & 0.188 & 0.280 & 0.383 & {\color{red}0.414} & {\color{blue}0.401} & {\color{green}0.408}    \\
Accuracy    & 0.418 & 0.463 & 0.530 & 0.490 & 0.494 & 0.503 & 0.484 & {\color{blue}0.586} & {\color{red}0.600} & {\color{green}0.590} & {\color{blue}0.586}    \\
Robustness  & 1.297 & 0.534 & 0.688 & 0.974 & 0.318 & 0.585 & {\color{blue}0.276} & {\color{blue}0.276} & {\color{green}0.234} & {\color{red}0.204}  & 0.281    \\
\hline
\end{tabular}
}
\end{table}
\begin{table}[!h]
\caption{Evaluation on LaSOT with protocol I (testing on all videos) and protocol II (training on given videos and testing on the rest). We get better results with dataset's own videos as training due to lare training set and matching domain.}
\label{tab:eval_lasot}
\centering
\scalebox{0.7}{
\renewcommand{\arraystretch}{1.1}
\begin{tabular}{@{}l c c c c c c c c c c c c c c c c@{}}
\hline
& {\small STAPLE} & {\small SRDCF} & {\small SiamFC}& {\small SINT} & {\small MDNet}& {\small ECO}& {\small BACF}& {\small VITAL}& {\small ATOM} &{\small SiamRPN++}&{\small DiMP}& {\small \textbf{Ours}} \\ \hline
(I) Accuracy     & 0.266 & 0.271 & 0.358 & 0.339 & {\color{red}0.413} & 0.340 & 0.277 & 0.412 & {\color{blue}0.515}&0.496&{\color{red}0.596}&{\color{green}0.521} \\
(I) Robustness  & 0.231 & 0.227 & 0.341 & 0.229 & {\color{green}0.374} & 0.298 & 0.239 & {\color{blue}0.372} &-&-&-&{\color{red}0.411} \\
\hline
(II)Accuracy    & 0.243 & 0.245 & 0.336 & 0.314 & {\color{green}0.397} & 0.324 & 0.259 & {\color{blue}0.390} &-&-&-&{\color{red}0.507}  \\
(II)Robustness  & 0.239 & 0.219 & 0.339 & 0.295 & {\color{green}0.373} & 0.301 & 0.239 & {\color{blue}0.360} &-&-&-&{\color{red}0.499} \\ 
\hline
\end{tabular}
}
\end{table}
\begin{table}[!h]
\caption{Evaluation on UAV123 by success rate and precision. Our algorithm is having difficulty with small/ low resolution targets.}
\label{tab:eval_uav123}
\centering
\scalebox{0.7}{
\renewcommand{\arraystretch}{1.1}
\begin{tabular}{@{}l c c c c c c c c c c c c c c@{}}
\hline
& {\small TLD} & {\small STRUCK}& {\small MEEM} & {\small STAPLE} & {\small SRDCF} & {\small MUSTer} & {\small ECO} & {\small ATOM} & {\small SiamRPN}& {\small SiamRPN++}& {\small DiMP} & {\small \textbf{Ours}} \\ \hline
Success   & 0.283 & 0.387 & 0.398 & 0.453 & 0.473 & 0.517 & 0.399 & {\color{blue}0.650} & 0.527 & 0.613 & {\color{red}0.653}&{\color{green}0.651}\\
Precision & 0.439 & 0.578 & 0.627 & -     & 0.676 & -     & 0.591 & - & {\color{blue}0.748} & {\color{green}0.807} &-  & {\color{red}0.833}\\
\hline
\end{tabular}
}
\end{table}

\begin{table}[!h]
\caption{
Benchmarking on TrackingNet  and GOT-10k
}
\small
\center
\scalebox{0.75}{
\begin{tabular}{@{}ll@{}c@{}c@{}c@{}c@{}c@{}c@{}c@{}c@{}c@{}c@{}}
\hline
& & ECO & DaSiam-RPN & ATOM & SiamRPN++ & DiMP & SiamMask & D3S & SiamFC++ & SiamRCNN & ours  \\ 
\hline 
\parbox[t]{2mm}{\multirow{3}{*}{\rotatebox[origin=c]{90}{T-Net}}}& Prec. & 0.492 & 0.591 & 0.648 & 0.694 & 0.687 & {\color{blue}0.733} & - & 0.705 & {\color{red}0.800} & {\color{green}0.711}  \\ 
 & N-Prec. & 0.618 & 0.733 & 0.771 & 0.800 & {\color{blue}0.801} & 0.664 & - & 0.800 & {\color{red}0.854} & {\color{green}0.810}  \\ 
& Success & 0.554 & 0.638 & 0.703 & 0.733 & 0.740 & {\color{green}0.778} & - & {\color{blue}0.754} & {\color{red}0.812} & 0.752  \\ \hline
\parbox[t]{2mm}{\multirow{3}{*}{\rotatebox[origin=c]{90}{GOT10k}}} & AO & 0.316 & 0.417 & 0.556 & 0.518 & {\color{red}0.611} & 0.514 & {\color{blue}0.597} & 0.595 & - & {\color{green}0.601}  \\ 
& SR 0.75 & 0.111 & 0.149 & 0.402 & 0.325 & {\color{red}0.492} & 0.366 & {\color{blue}0.462} & {\color{green}0.479} & - & {\color{green}0.479}  \\ 
& SR 0.5 & 0.309 & 0.461 & 0.635 & 0.618 & {\color{red}0.717} & 0.587 & 0.676 & {\color{green}0.695} & - & {\color{blue}0.685}  \\ 
\hline
\end{tabular}}
\end{table}
\noindent\textbf{Recent Public Datasets:}
Our method is compared with recent state-of-the-art methods in VOT2018 \cite{kristan2018sixth}, UAV123 \cite{mueller2016benchmark}, LaSOT \cite{fan2019lasot}, GOT-10K\cite{huang2019got} and TrackingNet\cite{muller2018trackingnet} datasets. In phase I of LaSOT evaluation, our tracker is trained on our own data and tested on all 1400 training video sequences of LaSOT. In phase II, the training data is limited to the given 1120 training videos and tested on the rest.
The results are better than SiamRPN++ in VOT2018 and UAV123, and VITAL in LaSOT, despite using a pre-trained CNN. This method benefits from multi-layer fusion, adaptive model update, and various regularization.  
Comparing the results of the benchmark with ATOM, SiamRPN++ and DIMP showed that just using convolutional layers and using the underlying features is not enough to perform a high-level tracking. Having a deeper network (ResNet-18 in ATOM and ResNet-50 in DiMP) and having auxiliary branches (region proposal in SiamRPN, IOU prediction in ATOM, and model prediction in DIMP) are two main differences between our method and the SotA. However, the proposed method offers insight about extra features to be used in tracking, along with the activation such as cooccurrence features that are embedded in all CNNs ready to be exploited.
Further, good performance on specific datasets such as UAV123 demonstrates that these features can support different object types and contexts.

\section{Conclusion}
We proposed a tracker that exploits various CNN statistics including activations, spatial data, co-occurrences within a layer, and temporal changes and patterns between time slices. It adaptively fuses several CNN layers to negate the demerits of each layer with merits of others. It outperformed recent trackers in various experiments, promoting the use of spatial and temporal style in tracking. Our regularizations can be used with other CNN-based methods. 

\bibliographystyle{splncs}
\bibliography{refs}

\begin{thebibliography}{10}

\bibitem{DEEPTRACK}
Li, H., Li, Y., Porikli, F.:
\newblock Deeptrack: Learning discriminative feature representations online for
  robust visual tracking.
\newblock IEEE TIP \textbf{25} (2016)  1834--1848

\bibitem{DeepSRDCF}
Danelljan, M., Hager, G., Shahbaz~Khan, F., Felsberg, M.:
\newblock Convolutional features for correlation filter based visual tracking.
\newblock In: Proceedings of the IEEE International Conference on Computer
  Vision Workshops. (2015)  58--66

\bibitem{wang2015video}
Wang, L., Liu, T., Wang, G., Chan, K.L., Yang, Q.:
\newblock Video tracking using learned hierarchical features.
\newblock IEEE TIP \textbf{24} (2015)  1424--1435

\bibitem{zhang2017multi}
Zhang, T., Xu, C., Yang, M.H.:
\newblock Multi-task correlation particle filter for robust object tracking.
\newblock In: CVPR. Volume~1. (2017) ~3

\bibitem{CNN-SVM}
Hong, S., You, T., Kwak, S., Han, B.:
\newblock Online tracking by learning discriminative saliency map with
  convolutional neural network.
\newblock In: International Conference on Machine Learning. (2015)  597--606

\bibitem{sharif2014cnn}
Sharif~Razavian, A., Azizpour, H., Sullivan, J., Carlsson, S.:
\newblock Cnn features off-the-shelf: an astounding baseline for recognition.
\newblock In: CVPRw. (2014)  806--813

\bibitem{cimpoi2014deep}
Cimpoi, M., Maji, S., Vedaldi, A.:
\newblock Deep convolutional filter banks for texture recognition and
  segmentation.
\newblock arXiv preprint arXiv:1411.6836 (2014)

\bibitem{danelljan2015learning}
Danelljan, M., Hager, G., Shahbaz~Khan, F., Felsberg, M.:
\newblock Learning spatially regularized correlation filters for visual
  tracking.
\newblock In: ICCV'15. (2015)  4310--4318

\bibitem{TCNN}
Nam, H., Baek, M., Han, B.:
\newblock Modeling and propagating cnns in a tree structure for visual
  tracking.
\newblock arXiv preprint arXiv:1608.07242 (2016)

\bibitem{DLT}
Wang, N., Yeung, D.Y.:
\newblock Learning a deep compact image representation for visual tracking.
\newblock In: NIPS. (2013)  809--817

\bibitem{AE-ENS}
Zhou, X., Xie, L., Zhang, P., Zhang, Y.:
\newblock An ensemble of deep neural networks for object tracking.
\newblock In: Image Processing (ICIP), 2014 IEEE International Conference on,
  IEEE (2014)  843--847

\bibitem{fan2010human}
Fan, J., Xu, W., Wu, Y., Gong, Y.:
\newblock Human tracking using convolutional neural networks.
\newblock IEEE Transactions on Neural Networks \textbf{21} (2010)  1610--1623

\bibitem{CF2}
Ma, C., Huang, J.B., Yang, X., Yang, M.H.:
\newblock Hierarchical convolutional features for visual tracking.
\newblock In: ICCV. (2015)  3074--3082

\bibitem{CNT}
Zhang, K., Liu, Q., Wu, Y., Yang, M.:
\newblock Robust visual tracking via convolutional networks without training.
\newblock IEEE TIP \textbf{25} (2016)  1779--1792

\bibitem{UCT}
Zhu, Z., Huang, G., Zou, W., Du, D., Huang, C.:
\newblock Uct: learning unified convolutional networks for real-time visual
  tracking.
\newblock In: ICCVw. (2017)  1973--1982

\bibitem{YCNN}
Chen, K., Tao, W.:
\newblock Once for all: a two-flow convolutional neural network for visual
  tracking.
\newblock IEEE CSVT (2018)  1--1

\bibitem{SO-DLT}
Wang, N., Li, S., Gupta, A., Yeung, D.Y.:
\newblock Transferring rich feature hierarchies for robust visual tracking.
\newblock arXiv preprint arXiv:1501.04587 (2015)

\bibitem{drayer2016object}
Drayer, B., Brox, T.:
\newblock Object detection, tracking, and motion segmentation for object-level
  video segmentation.
\newblock arXiv preprint arXiv:1608.03066 (2016)

\bibitem{SINT}
Tao, R., Gavves, E., Smeulders, A.W.:
\newblock Siamese instance search for tracking.
\newblock In: CVPR. (2016)  1420--1429

\bibitem{SIAMESEFC}
Bertinetto, L., Valmadre, J., Henriques, J.F., Vedaldi, A., Torr, P.H.:
\newblock Fully-convolutional siamese networks for object tracking.
\newblock In: ECCV, Springer (2016)  850--865

\bibitem{li2018high}
Li, B., Yan, J., Wu, W., Zhu, Z., Hu, X.:
\newblock High performance visual tracking with siamese region proposal
  network.
\newblock In: CVPR'18. (2018)  8971--8980

\bibitem{li2019siamrpn++}
Li, B., Wu, W., Wang, Q., Zhang, F., Xing, J., Yan, J.:
\newblock Siamrpn++: Evolution of siamese visual tracking with very deep
  networks.
\newblock In: CVPR'19. (2019)  4282--4291

\bibitem{SINTpp}
Wang, X., Li, C., Luo, B., Tang, J.:
\newblock Sint++: Robust visual tracking via adversarial positive instance
  generation.
\newblock In: Proceedings of the IEEE Conference on Computer Vision and Pattern
  Recognition. (2018)  4864--4873

\bibitem{VITAL}
Song, Y., Ma, C., Wu, X., Gong, L., Bao, L., Zuo, W., Shen, C., Rynson, L.,
  Yang, M.H.:
\newblock Vital: Visual tracking via adversarial learning.
\newblock In: CVPR. (2018)

\bibitem{Coloring}
Vondrick, C., Shrivastava, A., Fathi, A., Guadarrama, S., Murphy, K.:
\newblock Tracking emerges by colorizing videos.
\newblock In: ECCV. (2018)

\bibitem{zeiler2014visualizing}
Zeiler, M.D., Fergus, R.:
\newblock Visualizing and understanding convolutional networks.
\newblock In: ECCV, Springer (2014)  818--833

\bibitem{liu2015treasure}
Liu, L., Shen, C., van~den Hengel, A.:
\newblock The treasure beneath convolutional layers: Cross-convolutional-layer
  pooling for image classification.
\newblock In: CVPR. (2015)  4749--4757

\bibitem{HDTstar}
Qi, Y., Zhang, S., Qin, L., Huang, Q., Yao, H., Lim, J., Yang, M.H.:
\newblock Hedging deep features for visual tracking.
\newblock PAMI (2018)

\bibitem{CCOT}
Danelljan, M., Robinson, A., Khan, F.S., Felsberg, M.:
\newblock Beyond correlation filters: Learning continuous convolution operators
  for visual tracking.
\newblock In: ECCV, Springer (2016)  472--488

\bibitem{bovik1990multichannel}
Bovik, A.C., Clark, M., Geisler, W.S.:
\newblock Multichannel texture analysis using localized spatial filters.
\newblock PAMI (1990)  55--73

\bibitem{gatys2016image}
Gatys, L.A., Ecker, A.S., Bethge, M.:
\newblock Image style transfer using convolutional neural networks.
\newblock In: CVPR. (2016)  2414--2423

\bibitem{johnson2016perceptual}
Johnson, J., Alahi, A., Fei-Fei, L.:
\newblock Perceptual losses for real-time style transfer and super-resolution.
\newblock In: ECCV, Springer (2016)  694--711

\bibitem{matsuo2016cnn}
Matsuo, S., Yanai, K.:
\newblock Cnn-based style vector for style image retrieval.
\newblock In: Proceedings of the 2016 ACM on International Conference on
  Multimedia Retrieval, ACM (2016)  309--312

\bibitem{varol2018long}
Varol, G., Laptev, I., Schmid, C.:
\newblock Long-term temporal convolutions for action recognition.
\newblock PAMI \textbf{40} (2018)  1510--1517

\bibitem{gkioxari2015finding}
Gkioxari, G., Malik, J.:
\newblock Finding action tubes.
\newblock In: CVPR. (2015)  759--768

\bibitem{chao2018rethinking}
Chao, Y.W., Vijayanarasimhan, S., Seybold, B., Ross, D.A., Deng, J.,
  Sukthankar, R.:
\newblock Rethinking the faster r-cnn architecture for temporal action
  localization.
\newblock In: CVPR. (2018)  1130--1139

\bibitem{simonyan2014two}
Simonyan, K., Zisserman, A.:
\newblock Two-stream convolutional networks for action recognition in videos.
\newblock In: NIPS. (2014)  568--576

\bibitem{zhu2017end}
Zhu, Z., Wu, W., Zou, W., Yan, J.:
\newblock End-to-end flow correlation tracking with spatial-temporal attention.
\newblock CVPR \textbf{42} (2017) ~20

\bibitem{dosovitskiy2015flownet}
Dosovitskiy, A., Fischer, P., Ilg, E., Hausser, P., Hazirbas, C., Golkov, V.,
  Van Der~Smagt, P., Cremers, D., Brox, T.:
\newblock Flownet: Learning optical flow with convolutional networks.
\newblock In: Proceedings of the IEEE International Conference on Computer
  Vision. (2015)  2758--2766

\bibitem{feichtenhofer2018have}
Feichtenhofer, C., Pinz, A., Wildes, R.P., Zisserman, A.:
\newblock What have we learned from deep representations for action
  recognition?
\newblock connections \textbf{19} (2018) ~29

\bibitem{gladh2016deep}
Gladh, S., Danelljan, M., Khan, F.S., Felsberg, M.:
\newblock Deep motion features for visual tracking.
\newblock In: ICPR, IEEE (2016)  1243--1248

\bibitem{zhu2019stresnet_cf}
Zhu, Z.,  et~al.:
\newblock {STR}es{N}et\_cf tracker: The deep spatiotemporal features learning
  for correlation filter based robust visual object tracking.
\newblock IEEE Access \textbf{7} (2019)

\bibitem{danelljan2016beyond}
Danelljan, M., Robinson, A., Khan, F.S., Felsberg, M.:
\newblock Beyond correlation filters: Learning continuous convolution operators
  for visual tracking.
\newblock In: European Conference on Computer Vision, Springer (2016)  472--488

\bibitem{danelljan2017eco}
Danelljan, M., Bhat, G., Khan, F.S., Felsberg, M.:
\newblock Eco: Efficient convolution operators for tracking.
\newblock In: CVPR. (2017)

\bibitem{galoogahi2017learning}
Galoogahi, H.K., Fagg, A., Lucey, S.:
\newblock Learning background-aware correlation filters for visual tracking.
\newblock arXiv preprint arXiv:1703.04590 (2017)

\bibitem{danelljan2014accurate}
Danelljan, M., H{\"a}ger, G., Khan, F., Felsberg, M.:
\newblock Accurate scale estimation for robust visual tracking.
\newblock In: BMVC, BMVA Press (2014)

\bibitem{henriques2012exploiting}
Henriques, J.F., Caseiro, R., Martins, P., Batista, J.:
\newblock Exploiting the circulant structure of tracking-by-detection with
  kernels.
\newblock In: ECCV'12, Springer (2012)  702--715

\bibitem{bolme2010visual}
Bolme, D.S., Beveridge, J.R., Draper, B.A., Lui, Y.M.:
\newblock Visual object tracking using adaptive correlation filters.
\newblock In: CVPR'10, IEEE (2010)  2544--2550

\bibitem{jepson2003robust}
Jepson, A.D., Fleet, D.J., El-Maraghi, T.F.:
\newblock Robust online appearance models for visual tracking.
\newblock PAMI (2003)

\bibitem{berger2016incorporating}
Berger, G., Memisevic, R.:
\newblock Incorporating long-range consistency in cnn-based texture generation.
\newblock ICLR (2017)

\bibitem{wiyatno2019physical}
Wiyatno, R.R., Xu, A.:
\newblock Physical adversarial textures that fool visual object tracking.
\newblock In: Proceedings of the IEEE International Conference on Computer
  Vision. (2019)  4822--4831

\bibitem{vedaldi2015matconvnet}
Vedaldi, A., Lenc, K.:
\newblock Matconvnet: Convolutional neural networks for matlab.
\newblock In: Proceedings of the 23rd ACM international conference on
  Multimedia, ACM (2015)  689--692

\bibitem{real2017youtube}
Real, E., Shlens, J., Mazzocchi, S., Pan, X., Vanhoucke, V.:
\newblock Youtube-boundingboxes: A large high-precision human-annotated data
  set for object detection in video.
\newblock In: CVPR'17. (2017)  5296--5305

\bibitem{wu2013online}
Wu, Y., Lim, J., Yang, M.H.:
\newblock Online object tracking: A benchmark.
\newblock In: CVPR'13, IEEE (2013)  2411--2418

\bibitem{wu2015object}
Wu, Y., Lim, J., Yang, M.H.:
\newblock Object tracking benchmark.
\newblock PAMI (2015)

\bibitem{mueller2016benchmark}
Mueller, M.,  et~al.:
\newblock A benchmark and simulator for uav tracking.
\newblock In: ECCV'16, Springer (2016)

\bibitem{fan2019lasot}
Fan, H.,  et~al.:
\newblock La{SOT}: A high-quality benchmark for large-scale single object
  tracking.
\newblock CVPR'19 (2019)

\bibitem{kristan2015visual}
Kristan, M., Matas, J., Leonardis, A., Felsberg, M., Cehovin, L.,
  Fern{\'a}ndez, G., Vojir, T., Hager, G., Nebehay, G., Pflugfelder, R.:
\newblock The visual object tracking vot2015 challenge results.
\newblock In: ICCVw'15. (2015)  1--23

\bibitem{kristan2018sixth}
Kristan, M.,  et~al.:
\newblock The sixth visual object tracking vot2018 challenge results.
\newblock In: ECCV'18. (2018)

\bibitem{choi20123d}
Choi, C., Christensen, H.I.:
\newblock 3d textureless object detection and tracking: An edge-based approach.
\newblock In: 2012 IEEE/RSJ International Conference on Intelligent Robots and
  Systems, IEEE (2012)  3877--3884

\bibitem{kalal2012tracking}
Kalal, Z., Mikolajczyk, K., Matas, J.:
\newblock Tracking-learning-detection.
\newblock PAMI \textbf{34} (2012)  1409--1422

\bibitem{hare2011struck}
Hare, S., Saffari, A., Torr, P.H.:
\newblock Struck: Structured output tracking with kernels.
\newblock In: ICCV'11. (2011)

\bibitem{zhang2014meem}
Zhang, J., Ma, S., Sclaroff, S.:
\newblock Meem: Robust tracking via multiple experts using entropy
  minimization.
\newblock In: ECCV.
\newblock (2014)

\bibitem{hong2015multi}
Hong, Z., Chen, Z., Wang, C., Mei, X., Prokhorov, D., Tao, D.:
\newblock Multi-store tracker (muster): A cognitive psychology inspired
  approach to object tracking.
\newblock In: CVPR'15. (2015)  749--758

\bibitem{bertinetto2016staple}
Bertinetto, L., Valmadre, J., Golodetz, S., Miksik, O., Torr, P.H.:
\newblock Staple: Complementary learners for real-time tracking.
\newblock In: CVPR. (2016)  1401--1409

\bibitem{meshgi2017active}
Meshgi, K., Oba, S., Ishii, S.:
\newblock Active discriminative tracking using collective memory.
\newblock  (In: MVA'17)

\bibitem{danelljan2015convolutional}
Danelljan, M., Hager, G., Shahbaz~Khan, F., Felsberg, M.:
\newblock Convolutional features for correlation filter based visual tracking.
\newblock In: ICCVw. (2015)  58--66

\bibitem{danelljan2019atom}
Danelljan, M., Bhat, G., Khan, F.S., Felsberg, M.:
\newblock Atom: Accurate tracking by overlap maximization.
\newblock In: CVPR'19. (2019)  4660--4669

\bibitem{HDT}
Qi, Y., Zhang, S., Qin, L., Yao, H., Huang, Q., Lim, J., Yang, M.H.:
\newblock Hedged deep tracking.
\newblock In: CVPR. (2016)  4303--4311

\bibitem{MDNet}
Nam, H., Han, B.:
\newblock Learning multi-domain convolutional neural networks for visual
  tracking.
\newblock In: CVPR. (2016)  4293--4302

\bibitem{li2018learning}
Li, F.,  et~al.:
\newblock Learning spatial-temporal regularized correlation filters for visual
  tracking.
\newblock In: CVPR'18. (2018)

\bibitem{zhou2018efficient}
Zhou, Y.,  et~al.:
\newblock Efficient correlation tracking via center-biased spatial
  regularization.
\newblock IEEE TIP \textbf{27} (2018)  6159--6173

\bibitem{bhat2019learning}
Bhat, G., Danelljan, M., Gool, L.V., Timofte, R.:
\newblock Learning discriminative model prediction for tracking.
\newblock In: Proceedings of the IEEE International Conference on Computer
  Vision. (2019)  6182--6191

\bibitem{huang2019got}
Huang, L., Zhao, X., Huang, K.:
\newblock Got-10k: A large high-diversity benchmark for generic object tracking
  in the wild.
\newblock IEEE TPAMI (2019)

\bibitem{muller2018trackingnet}
Muller, M., Bibi, A., Giancola, S., Alsubaihi, S., Ghanem, B.:
\newblock Trackingnet: A large-scale dataset and benchmark for object tracking
  in the wild.
\newblock In: ECCV'2018. (2018)

\end{thebibliography}

\end{document}